\documentclass[journal]{IEEEtran}
\usepackage{amsmath}
\usepackage{amssymb}
\usepackage{multirow}

\usepackage{makecell}
\usepackage[colorlinks=true,linkcolor=blue,citecolor=blue,urlcolor=blue]{hyperref}

\usepackage[capitalise]{cleveref}
\usepackage{graphicx}

\usepackage{stfloats}

\usepackage{booktabs} 

\begin{document}

\title{DeTracker: Motion-decoupled Vehicle Detection and Tracking in Unstabilized Satellite Videos}

\author{Jiajun Chen,
        Jing Xiao,~\IEEEmembership{Senior~Member,~IEEE},
        Shaohan Cao,
        Yuming Zhu,
        Liang Liao,~\IEEEmembership{Senior~Member,~IEEE}, \\
        Jun Pan,~\IEEEmembership{Member,~IEEE}, 
        and Mi Wang,~\IEEEmembership{Member,~IEEE}
\thanks{This work was supported by the National Natural Science Foundation of China (62442107).}
\thanks{J. Chen, Y. Zhu, J. Pan and M. Wang are with the State Key Laboratory of Information Engineering in Surveying, Mapping and Remote Sensing, Wuhan University, Wuhan, China (e-mail: jiajunchen@whu.edu.cn; whuzym@whu.edu.cn; panjun1215@whu.edu.cn; wangmi@whu.edu.cn).}
\thanks{J. Xiao and S. Cao are with the School of Artificial Intelligence, Wuhan University, Wuhan, China (e-mail: jing@whu.edu.cn; csh0760@whu.edu.cn).}
\thanks{L. Liao is with Hangzhou Institute of Technology, Xidian University, Hangzhou, China (e-mail: liaoliang01@xidian.edu.cn).}}

\markboth{Submitted to IEEE Transactions on Geoscience and Remote Sensing}%
{Shell \MakeLowercase{\textit{et al.}}: Bare Demo of IEEEtran.cls for IEEE Journals}

\maketitle

\begin{abstract}
Satellite videos provide continuous observations of surface dynamics but pose significant challenges for multi-object tracking (MOT), especially under unstabilized conditions where platform jitter and the weak appearance of tiny objects jointly degrade tracking performance. To address this problem, we propose DeTracker, a joint-detection-and-tracking framework tailored for unstabilized satellite videos.
DeTracker introduces a task-driven Global--Local Motion Decoupling (GLMD) module to address the motion imbalance between dominant platform motion and weak target motion. It suppresses background-dominated motion via global semantic alignment at the feature level and captures target-specific motion through local refinement, improving trajectory stability and identity consistency.
In addition, a Temporal Dependency Feature Pyramid (TDFP) module is developed to perform cross-frame temporal feature fusion, enhancing the continuity and discriminability of tiny-object representations. We further construct a new benchmark dataset, \emph{SDM-Car-SU}, which simulates multi-directional and multi-speed platform motions to enable systematic evaluation of tracking robustness under varying motion perturbations. Extensive experiments on both simulated and real unstabilized satellite videos demonstrate that DeTracker significantly outperforms existing methods, achieving 61.1\% MOTA on \emph{SDM-Car-SU} and 45.3\% MOTA on real satellite video data.
The code and dataset will be publicly available at https://github.com/alex-chenjiajun/DeTracker.
\end{abstract}


\IEEEpeerreviewmaketitle

\section{Introduction}

\IEEEPARstart{O}{bject} tracking in satellite videos is an important and emerging technique for understanding dynamic processes on the Earth’s surface~~\cite{sti}. Compared with traditional static remote sensing imagery, satellite videos provide continuous spatio-temporal observations, which effectively support the long-term analysis of ground targets and demonstrate great potential in applications such as urban traffic monitoring~\cite{Zhu2019Transportation}, disaster emergency response~\cite{Sai2025Disaster}, and smart city management~\cite{holail2024time}. However, due to the high-altitude imaging nature of satellite platforms, targets in satellite videos are typically small in scale and densely distributed, posing significant challenges to multi-object tracking (MOT).

Early MOT approaches primarily follow the tracking-by-detection (TBD) paradigm~\cite{Henriques2015KCF,Danelljan2014DSST,Xiao2022DSFNet,Wu2024HRTracker,He2022TGraM}, in which object detectors are used to localize targets in each frame, followed by cross-frame association based on re-identification features. With continued progress, spatio-temporal context modeling~\cite{Zhao2024MP2Net,mmframe}, cross-scale feature fusion~\cite{tmmxiao,Chen2024GMFTracker}, and Transformer-based architectures~\cite{Zhu2024TabCtNet,Zheng2023Target-Aware} have been incorporated to better handle complex background disturbances and capture long-range dependencies across frames. More recently, joint-detection-and-tracking (JDT) frameworks have emerged as a dominant approach~\cite{He2018Twofold,Zhou2019Deep,Lv2024TrackingWithRobustID,He2024FSTrack}. They employ a unified backbone to predict object bounding boxes and identity embeddings across consecutive frames within a single forward pass, thereby performing both object localization and identity association jointly. By leveraging identity matching via embedding similarity, these methods reduce reliance on manually designed association rules and achieve more consistent MOT performance in large-scale dynamic scenes.

At present, most MOT methods for satellite videos are developed on pre-stabilized video sequences by video stabilization methods~\cite{Li2022Stabilization,Zhang2023High-Precision,Zhang2023Stabilization}, where background motion due to satellite platform movement is compensated to reduce its disturbances to object detection and improve the accuracy of object association.
Despite these advances, most existing satellite video tracking methods still rely on a stabilize-then-track paradigm. In this paradigm, the stabilization stage typically involves computationally expensive motion estimation and resampling operations, which makes it primarily suitable for offline ground-based processing after data downlink and difficult to meet the requirements of current online target information extraction and service applications.

Moreover, in addition to the limitations of existing processing paradigms, large-scale publicly available datasets for unstabilized satellite videos remain scarce. To comply with the stabilize-then-track pipeline, most existing satellite video MOT benchmarks are built on pre-stabilized videos~\cite{Zhang2020SkySat,Li2023SAT-MTB,Yin2022VISO,Zhang2024SDM-Car}, which limits the exploration of learning-based MOT methods in unstabilized scenarios.

However, directly transferring existing MOT methods designed for stabilized satellite videos to unstabilized satellite video scenarios often leads to a series of performance degradation issues, such as unstable feature matching and failure of target appearance modeling. This is mainly due to the complex disturbances introduced by satellite platform motion and attitude jitter in unstabilized satellite videos, which severely disrupt temporal consistency and background stability. The resulting challenges can be primarily summarized as the following two aspects:

\begin{figure*}[!htbp]
    \centering
    \includegraphics[width=\textwidth]{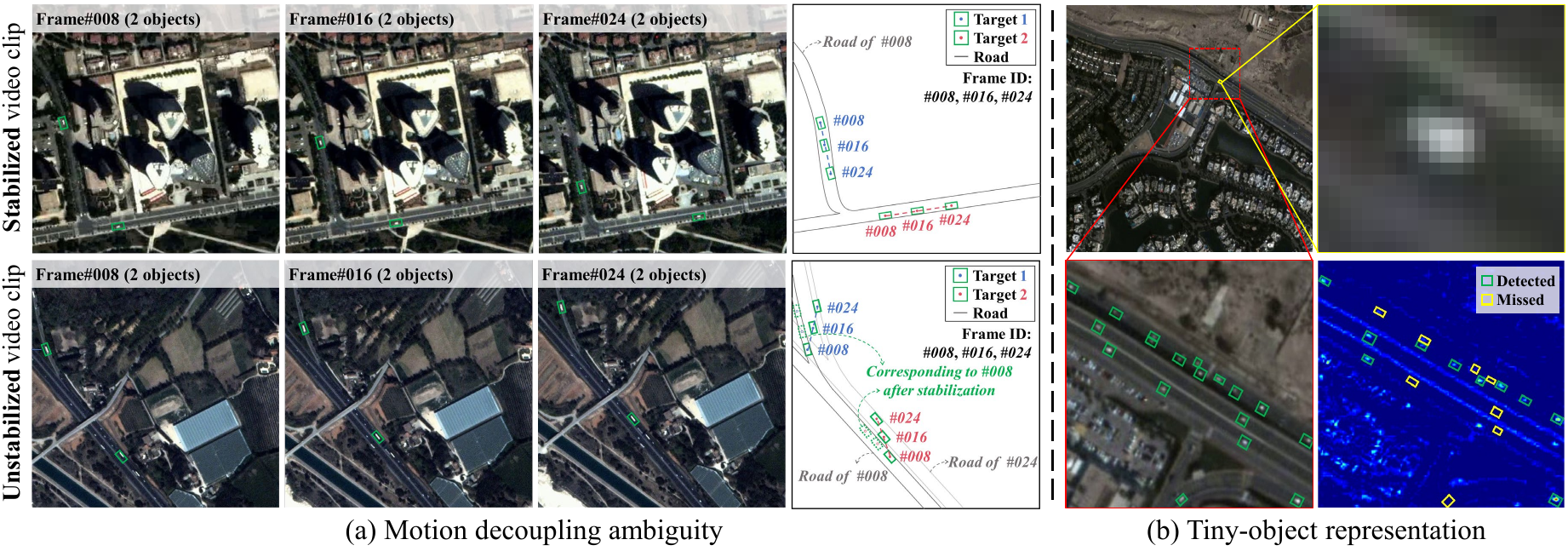}
    \vspace{-4mm}
    \caption{Major challenges of object tracking in unstabilized satellite videos. (a) Motion decoupling ambiguity: platform-induced jitter introduces complex global motion, causing discrepancies between the true trajectories of ground objects and the apparent motion observed in unstabilized satellite videos; (b) Tiny-object representation: objects of interest exhibit extremely small scales and limited visual cues, leading to insufficient feature representation.}
    \label{fig:Q1}
    \vspace{-3mm}
\end{figure*}

\subsubsection{Ambiguity in Decoupling of Mixed Motions} In unstabilized satellite videos, the observed motion arises from a superposition of true target motion and global motion induced by the satellite platform~\cite{Zhang2023Stabilization}. As illustrated in Fig.~\ref{fig:Q1}(a), even slight attitude perturbations can shift the sensor's field of view, making originally static ground backgrounds appear highly dynamic. Moreover, the magnitude of platform-induced motion is typically larger than the displacement of tiny targets, thereby dominating the observed motion. This coupling between platform motion and object motion makes motion decoupling inherently ambiguous, which in turn degrades motion estimation and leads to tracking trajectory drift.

\subsubsection{Difficulty in Feature Extraction of Tiny Objects} The field of view of satellite videos typically spans vast areas, while tiny targets often occupy only a few pixels, providing very limited texture and structural cues~\cite{Ren2024Motion}. In stabilized satellite videos, existing methods accumulate and enhance the features of tiny targets by modeling cross-frame correlations, thereby alleviating the insufficiency of single-frame features. However, in unstabilized satellite videos, satellite platform motion and attitude perturbations weaken the frame-to-frame consistency, making feature extraction and discriminative modeling of tiny targets more challenging. This significantly reduces feature distinctiveness and frequently leads to missed detections of tiny targets, as illustrated in Fig.~\ref{fig:Q1}(b).

To address the above challenges, we propose DeTracker, a joint-detection-and-tracking framework tailored for unstabilized satellite videos. Unlike conventional approaches that rely on explicit image-level stabilization, DeTracker performs background motion compensation directly in the feature space, enabling the network to learn motion-aware representations even under strong global motion disturbances.
Specifically, to address the motion coupling issue, we design a Global--Local Motion Decoupling (GLMD) module that explicitly decouples the global motion induced by the satellite platform from the target's intrinsic motion at the feature level, thereby suppressing background drift while preserving target-level motion cues critical for tracking. Within the GLMD module, a Global Alignment (GA) submodule leverages a motion-oriented self-attention mechanism to model long-range semantic dependencies across consecutive frames, focusing on suppressing platform-induced background feature variations to achieve cross-frame semantic-level alignment. Complementing the global alignment, a Local Refinement (LR) submodule learns region-level motion offsets in the preliminarily aligned feature space, capturing the fine-grained target motion information that was originally masked by background drift and thereby enhancing the discriminability of tiny objects.
Furthermore, we propose a Temporal Dependency Feature Pyramid (TDFP) module to improve feature extraction and modeling of tiny targets in unstabilized satellite videos, where frame-to-frame consistency is weakened. By performing multi-level temporal feature fusion and explicitly modeling cross-frame dependencies, TDFP enhances the spatio-temporal consistency of target representations, thereby improving detection robustness and tracking stability.

The main contributions are summarized as follows:
\begin{enumerate}

  \item We propose {DeTracker}, an MOT framework for small targets in unstabilized satellite videos, tailored for on-orbit information services. Experimental results show that DeTracker achieves MOTA improvements of 5.1\% and 8.3\% on the simulated dataset and real satellite videos, respectively, outperforming existing state-of-the-art methods.

  \item We propose the Global--Local Motion Decoupling (GLMD) module and the Temporal Dependency Feature Pyramid (TDFP) module. The GLMD explicitly decouples platform motion from target motion via feature-level global alignment and local refinement, effectively suppressing background drift for robust tracking. The TDFP models cross-frame dependencies and aggregates multi-scale features to enhance the temporal discriminability of small targets.

  \item To facilitate research on tracking in unstabilized satellite videos, we construct a \textbf{S}imulated \textbf{U}nstabilized satellite video MOT benchmark, \emph{SDM-Car-SU}, incorporating multi-directional and multi-velocity perturbations to emulate satellite platform motion. In addition, several real unstabilized satellite video clips are provided to evaluate algorithm performance in practical on-orbit scenarios.

\end{enumerate}

\section{Related Work}
\subsection{Object Tracking in Generic Video}
In general video object tracking, existing methods can be broadly divided into two paradigms: TBD and JDT. TBD methods perform frame-wise detection followed by cross-frame association using motion and appearance cues. DeepSORT~\cite{Wojke2017Simple} introduced learning-based appearance embeddings, laying the foundation for subsequent extensions such as StrongSORT~\cite{StrongSORT} and OC-SORT~\cite{cao2023OCsort}. ByteTrack~\cite{Bytetrack} uses both high-confidence and low-confidence detections in a two-stage association process, allowing low-confidence detections to recover objects during occlusion or weak detection. Recent studies further enhance association by integrating adaptive Kalman filtering with recurrent neural networks and hierarchical contextual modeling~\cite{Zhen2021Multiple,Li2024HCgNet,Li2024Single-Shot}.

The JDT paradigm unifies detection and association within a single network, typically achieving more stable identity consistency at the cost of higher training complexity and weaker cross-domain generalization.
Early JDT methods leverage Siamese architectures to strengthen the discriminative representation of object features~\cite{He2018Twofold,Zhou2019Deep}, while subsequent works enhance identity consistency and tracking robustness by incorporating visibility modulation, two-stage association, and feature enhancement, thereby reducing ID switches~\cite{Lv2024TrackingWithRobustID,He2024FSTrack}. More recently, Transformer-based JDT methods introduce learnable identity embeddings and data-driven association mechanisms, replacing heuristic matching with end-to-end identity prediction~\cite{Zhu2024TabCtNet,Zheng2023Target-Aware,Gao2025Multiple}.

Overall, despite their success in ground-based and camera-static scenarios, these general tracking frameworks face significant challenges when directly applied to satellite videos, motivating the need for adaptations to the unique motion patterns and imaging characteristics of satellite platforms.

\subsection{Object Tracking in Remote Sensing Video}
Research on object tracking in remote sensing video can be broadly divided into correlation filter-based methods and deep learning-based methods.
\subsubsection{Correlation Filter-based Methods}
Correlation filter-based methods are among the earliest techniques applied to UAV and satellite video tracking owing to their high computational efficiency. While they struggle under rapid and unstable UAV motion, the characteristics of satellite videos, such as tiny objects, low signal-to-noise ratios, complex backgrounds, and relatively slow motion, make correlation filters the dominant paradigm in the early and middle stages of satellite video tracking~\cite{Henriques2015KCF,Danelljan2014DSST}.
Subsequent works enhance robustness by integrating multiple filters~\cite{CFKF2019,Du2018Object}, incorporating motion estimation and multi-feature modeling into the KCF framework for low-resolution imagery~\cite{Zhang2022Satellite}, or combining Kalman filtering with temporal consistency constraints to mitigate drift under occlusion and background clutter~\cite{Ran2023Aircraft}.

\subsubsection{Deep Learning-based Methods}
Benefiting from flexible viewpoints and wide coverage, UAV videos have become a major testbed for deep learning-based tracking. 
Siamese network-based methods significantly improve tracking accuracy and robustness~\cite{Christoph2022ToMP,Christoph2021Keep,Gao2022AiATrack}, while scale-aware and motion-aware frameworks address object size variation and complex dynamics~\cite{hutcsvt,Li2024Edge,Zou2025MOSAIC}. Transformer-based architectures further enhance modeling capacity by jointly improving computational efficiency and capturing long-range dependencies~\cite{Cai2023Robust,Chen2023SeqTrack,Yang2023Foreground}. To cope with low-resolution inputs, recent works adopt knowledge distillation and feature compensation mechanisms to preserve performance while reducing computational cost~\cite{Li2024Learning,Yuan2025TLSH-MOT}.
In addition, several studies have explored spatio-temporal feature interaction mechanisms to alleviate frequent ID switches caused by small object scales, sparse feature representations, and significant appearance variations in UAV scenarios. By introducing spatio-temporal trajectory priors to enhance feature association, these methods effectively mitigate visual feature degradation and improve identity consistency~\cite{Wu2024TSMMT,Wu2025TCFNet}.

In satellite videos, objects are typically extremely small within the image, which makes feature extraction highly challenging and leads to severe feature ambiguity.
To alleviate this, researchers have proposed end-to-end frameworks with enhanced feature extractors to better represent blurred tiny objects~\cite{Xiao2022DSFNet,Wu2024HRTracker,He2022TGraM}. By incorporating cross-frame motion modeling and spatio-temporal fusion, these methods strengthen temporal continuity, while multi-scale cascaded enhancement modules further improve scale adaptability~\cite{WANG2024MCTracker}.
To further advance motion modeling, a motion-inspired cross-modality learning mechanism has been introduced~\cite{Chen2025MICPL}. By mining global motion flows to capture latent motion patterns, this approach bridges the semantic gap between motion and visual features, significantly improving the representation of dynamic tiny objects.
Other approaches exploit dynamic object masks to highlight tiny-object regions or employ sparse sampling strategies to correct feature relationships and suppress background interference~\cite{Zhao2024MP2Net}.
Building on the concept of sparsity, recent methods utilizing sparse spatio-temporal point cloud representations reduce redundant background clutter and enhance the detection of tiny, low-contrast objects~\cite{Xiao2024HiEUM}.
Recent works adopt dual-frame input with heatmap-based output and lightweight matching to preserve identity consistency, thereby enhancing the discriminability of tiny objects under dynamic noise~\cite{Kong2023CFTracker,Chen2024GMFTracker}. Trajectory-aware tracking frameworks further exploit motion history to improve both accuracy and efficiency~\cite{Chen2025PIFTrack,Chen2022Vehicle}.

In summary, although notable progress, satellite video tracking still poses substantial challenges arising from dense tiny objects, weak feature representations, and the strong coupling between platform motion and object trajectories, motivating the need for dedicated frameworks that explicitly decouple motion and enhance tiny-object representations.

\section{SDM-Car-SU: An Unstabilized Satellite Video Vehicle Tracking Dataset}

In satellite video MOT, datasets such as SkySat~\cite{Zhang2020SkySat}, VISO~\cite{Yin2022VISO}, SAT-MTB~\cite{Li2023SAT-MTB}, and SDM-Car~\cite{Zhang2024SDM-Car} have greatly advanced deep learning-based vehicle tracking methods, such as DSFNet~\cite{Xiao2022DSFNet} and MP2Net~\cite{Zhao2024MP2Net}. However, these publicly available datasets are primarily built from ground-preprocessed, stabilized videos, which differ significantly from raw on-orbit streams affected by platform jitter.

In practice, satellite orbital motion and attitude perturbations inevitably introduce global background motion, leading to inherently \emph{unstabilized} video sequences. To approximate such on-orbit imaging conditions, we construct a simulated unstabilized satellite video MOT dataset, \emph{SDM-Car-SU}, by extending SDM-Car~\cite{Zhang2024SDM-Car} with moving imaging windows. The dataset characteristics and construction procedure are described below.

\begin{figure*}[!htb]
    \centering
    \setlength{\tabcolsep}{1pt}
    {\small
    \begin{tabular}{c}
        \includegraphics[height=3cm,keepaspectratio]{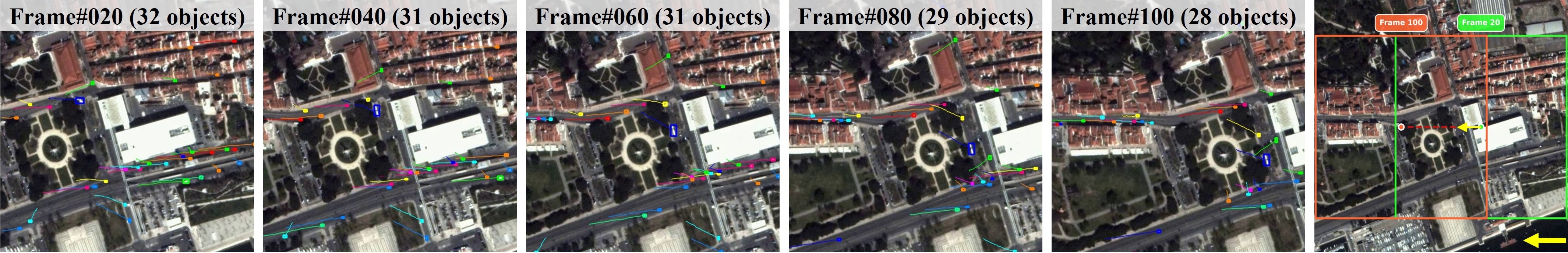} \\
        (a) An unstabilized sequence with lateral perturbations \\[1pt]

        \includegraphics[height=3cm,keepaspectratio]{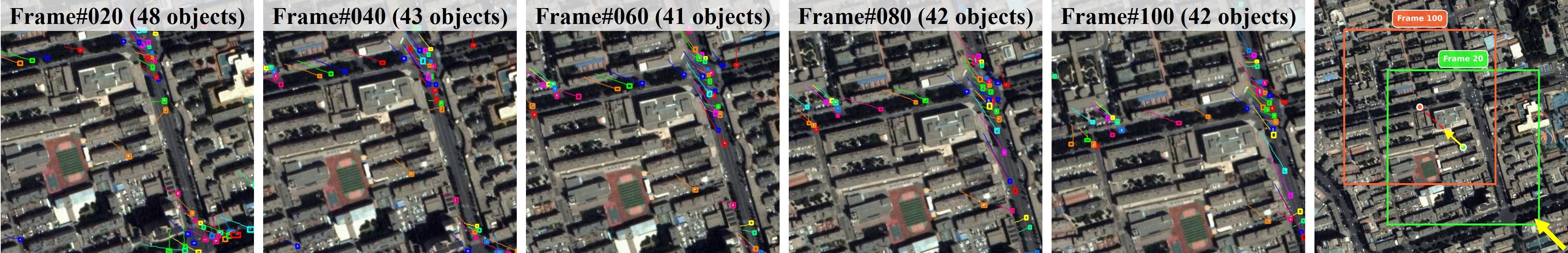} \\
       (b) An unstabilized sequence with diagonal perturbations \\[1pt]

        \includegraphics[height=3cm,keepaspectratio]{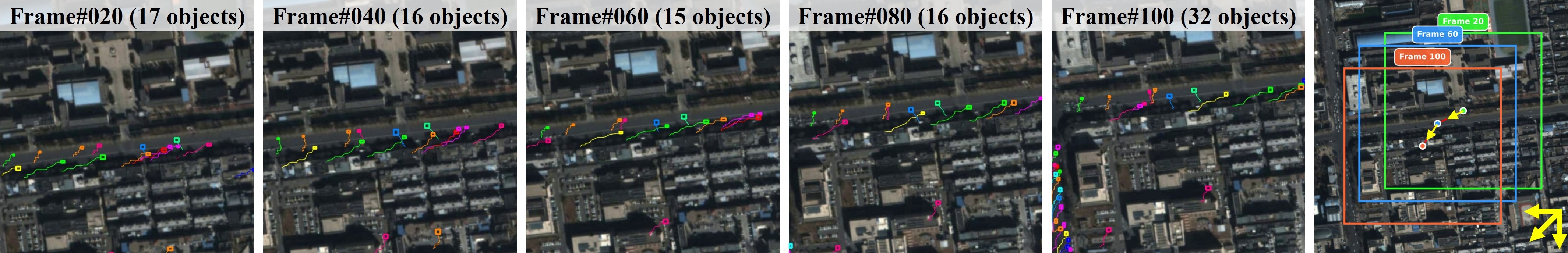} \\
        (c) An unstabilized sequence with mixed lateral and diagonal perturbations
    \end{tabular}
    }
    \vspace{-3mm}
    \caption{Visualization of inter-frame motion, annotation mapping, and target trajectories under unstabilized conditions. The left columns present the frame-wise annotation mapping results along with the coupled motion trajectories over the preceding 20 frames under different types of unstabilized perturbations, while the rightmost column visualizes the simulated platform motion directions.}
    \vspace{-3mm}
    \label{fig:dataset}
\end{figure*}

\subsection{Dataset Characteristics}
\emph{SDM-Car-SU} targets vehicle tracking in unstabilized satellite videos, as shown in Fig.~\ref{fig:dataset}. By simulating different motion directions, velocities, and combined motion patterns, it more faithfully reflects the complex background motion encountered in on-orbit imaging, thereby posing greater challenges for both detection and tracking.
Although DAF~\cite{Yu2025On-Satellite} considers simulated unstabilized data, it varies only background motion speed, neglects motion direction, and provides just one simulated sequence, which limits its applicability for training and benchmarking learning-based MOT models.
In contrast, \emph{SDM-Car-SU} offers a more comprehensive and diversified modeling of background motion along three dimensions:

\begin{itemize}
    \item \textbf{Direction dimension:} \emph{SDM-Car-SU} includes eight fundamental motion directions, consisting of up, down, left, right, and the four diagonal directions. For each direction, two different starting locations are defined as the initial positions of the motion, producing 16 unique start-point--direction combinations. This strategy ensures thorough coverage of different spatial regions in the original data.
    \item \textbf{Velocity dimension:} The dataset includes three motion speeds (1, 2, and 3 pixels per frame), spanning background disturbance levels from slower to faster than typical vehicle motion. This enables systematic evaluation of algorithm robustness under different jitter intensities.
     \item \textbf{Motion dimension:} For the test set, mixed motion patterns are designed. Since directly switching among arbitrary motion directions may lead to unrealistic relative motions, we define four combined motion patterns from the eight basic directions: 
    \{right, up, up-right\}, \{left, up, up-left\}, \{right, down, down-right\}, and \{left, down, down-left\}. Within each pattern, the three directions are randomly permuted to generate the test sequences.
\end{itemize}

Through this design, \emph{SDM-Car-SU} provides the simulated dataset specifically tailored for vehicle tracking in unstabilized satellite videos, providing a solid basis for assessing algorithm robustness and inductive biases under on-orbit conditions.

\subsection{Dataset Construction} 

The satellite data are sourced from the original SDM-Car dataset~\cite{Zhang2024SDM-Car}, which contains 99 high-quality video sequences acquired by the Luojia-3 01 satellite~\cite{Wang2024Luojia}, with a spatial resolution of 0.75 m and an image size of $1920 \times 1080$. By introducing inter-frame motion into the video frames to simulate satellite platform movement, we construct the \emph{SDM-Car-SU} dataset, which comprises three subsets with different motion rates: U1, U2, and U3, containing 335, 265, and 220 video clips, respectively. The train, validation, and test splits follow the same protocol as the SDM-Car dataset. 

In practical scenarios, due to the influence of platform disturbances, orbital and attitude measurement errors, as well as variations in viewing angles, adjacent frames inevitably exhibit residual translational and rotational deviations, thereby giving rise to characteristic unstabilized effects in satellite videos~\cite{Zhang2023Stabilization}. 
Theoretically, the geometric deviation of an image frame $I_{t+1}$ relative to a reference frame $I_t$ can be modeled by a 2D rigid transformation:  
\begin{equation}
X_{t+1} = R(\theta) X_t + d,
\end{equation}
where $X_t = [x_t, y_t]^T$ denotes the planar coordinates of a pixel at time $t$, $X_{t+1}$ represents the corresponding position of the pixel in the next frame, $\theta$ is the rotational deviation angle between adjacent frames, $R(\theta) = [\cos\theta, -\sin\theta; \sin\theta, \cos\theta]$ is the 2D rotation matrix, and $d = [d_x, d_y]^T$ denotes the translational deviation vector from time $t$ to $t+1$.

To facilitate variable control and emphasize the dominant geometric perturbations, we simplify the unstabilized imaging model to inter-frame translational deviations. By specifying various combinations of displacement directions and magnitudes, we construct 2D translational vectors $d = [d_x, d_y]^T$ to simulate the inter-frame shifts induced by orbital and attitude variations. Based on these vectors, a systematic simulated unstabilized satellite dataset is generated from stabilized data by applying a moving imaging window, thereby approximating realistic unstabilized imaging conditions.

Specifically, for motion parameter configuration, we follow the strategy of DAF~\cite{Yu2025On-Satellite} on the SkySat dataset~\cite{Zhang2020SkySat}, which has a spatial resolution of 1.0 m and background window speeds set to 0.5\textendash2.0 pixels per frame. These values are chosen according to the average vehicle motion at that resolution, thereby covering background perturbation levels from slower to faster than typical vehicle motion. Following the same rationale, we statistically analyze the inter-frame pixel displacements of the same vehicle targets in SDM-Car~\cite{Zhang2024SDM-Car} and obtain a mean motion speed of approximately 2 pixels per frame. Accordingly, we set the simulated background motion amplitudes to 1, 2, and 3 pixels per frame to create scenarios with different degrees of image instability.
Statistical analysis of 211 unstabilized satellite video clips from the Luojia-3 01 satellite validates the configured motion amplitudes. As demonstrated in Fig.~\ref{fig:hist}, approximately 90\% of the inter-frame motion magnitudes fall within a 3-pixel range. This empirical distribution provides robust substantiation for the chosen simulation parameters.
The construction process of the \emph{SDM-Car-SU} dataset is detailed as follows:
\begin{figure}[!t]
    \centering
    \includegraphics[width=0.49\textwidth]{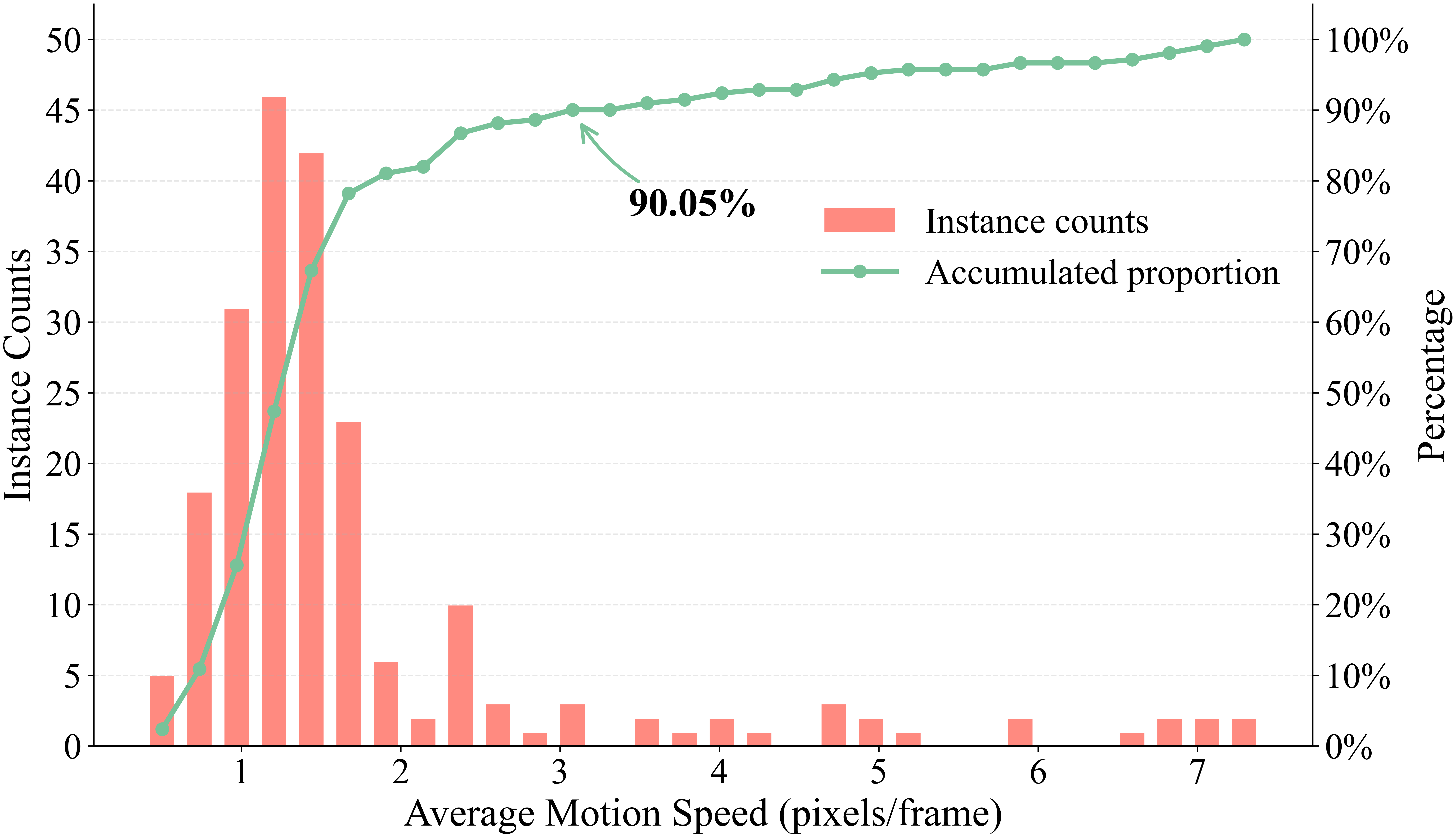}
    \caption{Distribution of inter-frame motion amplitudes in real-world unstabilized videos from the Luojia-3 01 satellite.}
    \label{fig:hist}
\end{figure}

\subsubsection{Moving Window Generation}
Let the center coordinates of the cropping window in the $n$-th frame be $(x_n, y_n)$, then its motion trajectory can be defined as:  
\begin{equation}
x_n = x_0 + v_x \cdot n, \quad y_n = y_0 + v_y \cdot n,
\end{equation}
where $(x_0, y_0)$ denotes the initial starting position, and $(v_x, v_y)$ represent the horizontal and vertical velocities (in pixels per frame). The velocity parameters are allowed to be floating-point values, thereby enabling sub-pixel motion simulation.

\subsubsection{Sub-pixel Precision Cropping}
For the floating-point coordinates $(x_n, y_n)$, we first extract the integer part $(\lfloor x_n \rfloor, \lfloor y_n \rfloor)$ for coarse cropping. Then, a translation matrix is constructed to compensate for the fractional offsets:
\begin{equation}
M = \begin{bmatrix}
1 & 0 & -(x_n - \lfloor x_n \rfloor) \\
0 & 1 & -(y_n - \lfloor y_n \rfloor)
\end{bmatrix}.
\end{equation}

Bilinear interpolation is subsequently applied to achieve sub-pixel-level smooth cropping, ensuring both continuity and precision in the motion sequence.  

\subsubsection{Annotation Filtering and Mapping}
To ensure the validity of the generated data, we retain only those sequences that satisfy the following criteria:  
(i) all cropped windows remain completely within the boundaries of the original image; and (ii) each frame contains at least one valid object annotation.

For an original bounding box $(x, y, w, h)$ and a cropping window of size $W_{\text{crop}} \times H_{\text{crop}}$, the new coordinates in the cropped frame are defined as:  
\begin{equation}
\begin{aligned}
x_{\text{new}} &= \max(x, x_n^{\text{crop}}) - x_n^{\text{crop}}, \\
y_{\text{new}} &= \max(y, y_n^{\text{crop}}) - y_n^{\text{crop}}, \\
w_{\text{new}} &= \min(x+w, x_n^{\text{crop}}+W_{\text{crop}}) - \max(x, x_n^{\text{crop}}), \\
h_{\text{new}} &= \min(y+h, y_n^{\text{crop}}+H_{\text{crop}}) - \max(y, y_n^{\text{crop}}),
\end{aligned}
\end{equation}
where $(x_n^{\text{crop}}, y_n^{\text{crop}})$ is the top-left coordinate of the cropping window, and $W_{\text{crop}}$ and $H_{\text{crop}}$ denote its width and height.

By applying the above procedure, we construct simulated unstabilized datasets with an image resolution of $512 \times 512$, providing a controlled testbed for evaluating tracking performance under complex and dynamically varying backgrounds.

\section{Proposed Method}
This section introduces the proposed DeTracker. We outline the tracking framework in Section \uppercase\expandafter{\romannumeral4}-A. Then, Section \uppercase\expandafter{\romannumeral4}-B presents the Global--Local Motion Decoupling (GLMD) module for motion disentanglement and Section \uppercase\expandafter{\romannumeral4}-C describes the Temporal Dependency Feature Pyramid (TDFP) module for temporal feature enhancement. The objective functions are detailed in Section \uppercase\expandafter{\romannumeral4}-D.

\subsection{Framework Overview}
\begin{figure*}[!h]
    \centering
    \includegraphics[width=\textwidth]{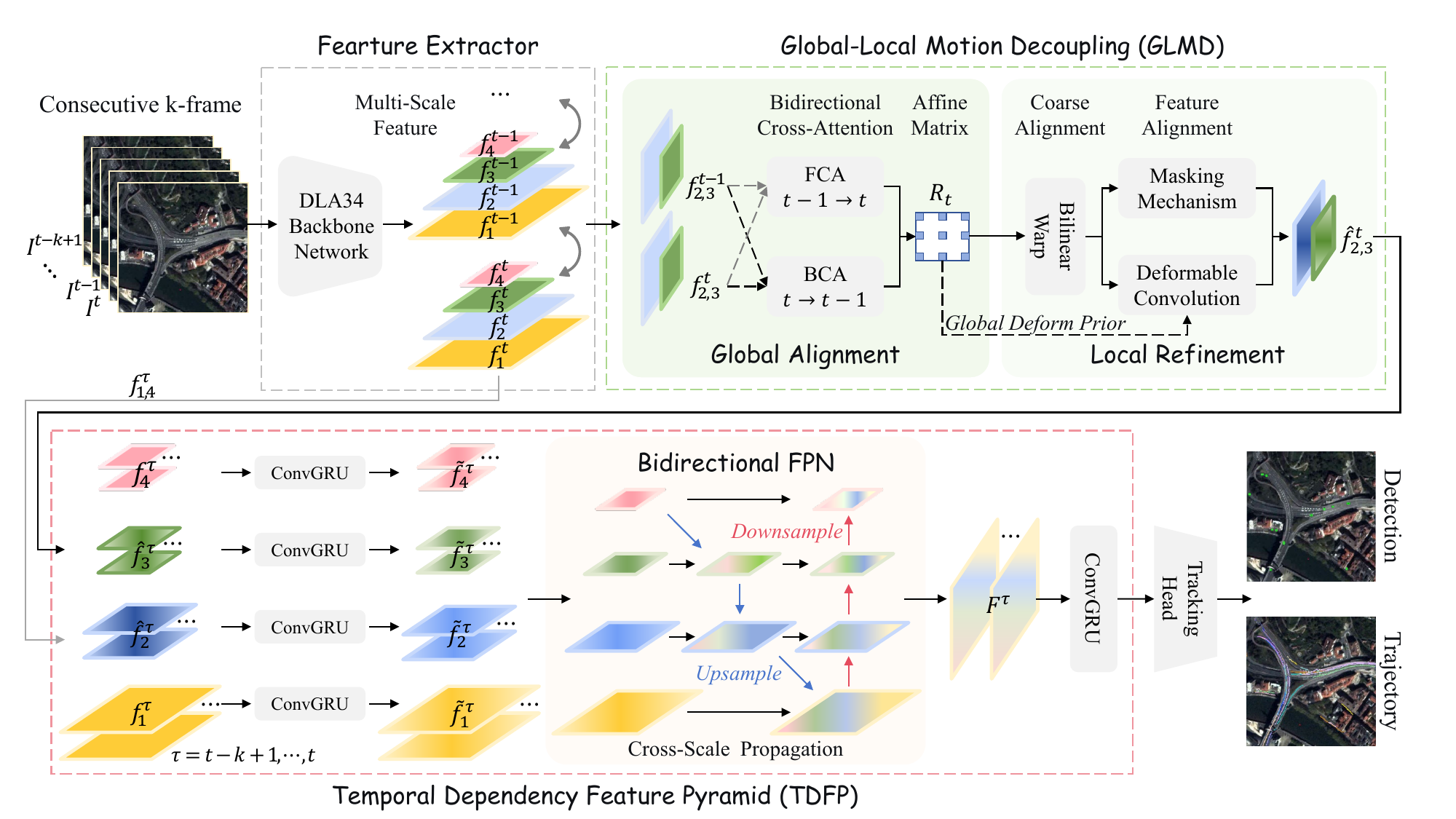}
    \caption{Overview of the proposed DeTracker. The framework consists of three components:
    (1) a feature extraction backbone for acquiring multi-scale spatial representations from input frames;
    (2) a Global–Local Motion Decoupling (GLMD) module for disentangling platform-induced global motion from target-level local motion;
    and (3) a Temporal Dependency Feature Pyramid (TDFP) module for aggregating temporal information and modeling cross-frame dependencies.}
    \label{fig:network}
\end{figure*}

Unlike conventional MOT pipelines that rely on stabilized inputs or post-hoc association, the proposed DeTracker directly operates on unstabilized satellite video and performs detection and tracking in an end-to-end manner. 
This design avoids hand-crafted motion compensation and enables the model to learn intrinsic motion structures directly from data, rather than relying on optical flow or Kalman-based heuristics.

As illustrated in Fig.~\ref{fig:network}, DeTracker consists of three components: a feature extraction backbone, a GLMD module, and a TDFP module. Specifically, given a video sequence $V=\{{I^n \mid n=1,2,\ldots,N}\}$, where $I^n$ is the $n$-th frame and $N$ is the total number of frames, we employ DLA34~\cite{DLA34_2018} as the backbone and take a contiguous clip of $k$ frames ending at time $t$, \emph{i.e.}, $V_{clip}=\{I^\tau \mid \tau=t-k+1,\ldots,t\}$, as input. For each input frame $I^\tau$, the backbone outputs a set of multi-scale feature maps $\{f_{i}^{\tau} \mid i = 1,2,3,4 \}$, where $i$ indexes the feature level from low to high dimension. 

Subsequently, mid-level features are selected for motion modeling, as shallow features susceptible to noise, while deep features
tend to collapse tiny objects into single pixels. The use of a stable $k$-frame input allows motion cues to accumulate in feature space instead of being reconstructed through explicit data association.
Specifically, the mid-level features $\{f^{\tau}_j\mid j=2,3\}$ from adjacent frames are fed into the GLMD module to correct misalignment in the current frame, where global translation trends and local fine-grained offsets are modeled, respectively. Together with low-level and high-level features, the motion-refined mid-level features form an updated feature set  $\{\hat{f}_i^{\tau} = (f_1^{\tau}, \hat{f}_2^{\tau}, \hat{f}_3^{\tau}, f_4^{\tau}) \mid \tau = t-k+1, \dots, t \}$.
Then, $\{\hat{f}_i^{\tau}\}$ is fed into the TDFP module, yielding a compact, low-dimensional feature set $\{F^{\tau}\mid \tau=t-k+1,\ldots,t\}$ enriched with semantic and temporal information. Finally, $\{F^{\tau}\}$ is fed into the detection head, where object detection and tracking are jointly performed via center-point prediction.

\subsection{Global--Local Motion Decoupling}

In this work, we decompose motion within unstabilized video sequences into two levels: global motion and local motion. Global motion arises from platform attitude changes and can be approximated by rigid, globally consistent transformations, while local motion remains after global compensation and encompasses non-rigid object dynamics alongside residual background errors. 
This hierarchical motion structure poses stringent requirements on feature learning: the network must capture long-range, cross-frame global dependencies while simultaneously preserving spatial precision and local detail consistency. We designed a feature-level motion compensation module that explicitly models temporal consistency, thereby stabilizing feature representations across consecutive frames.

\begin{figure}[!t]
    \centering
    \includegraphics[width=0.49\textwidth]{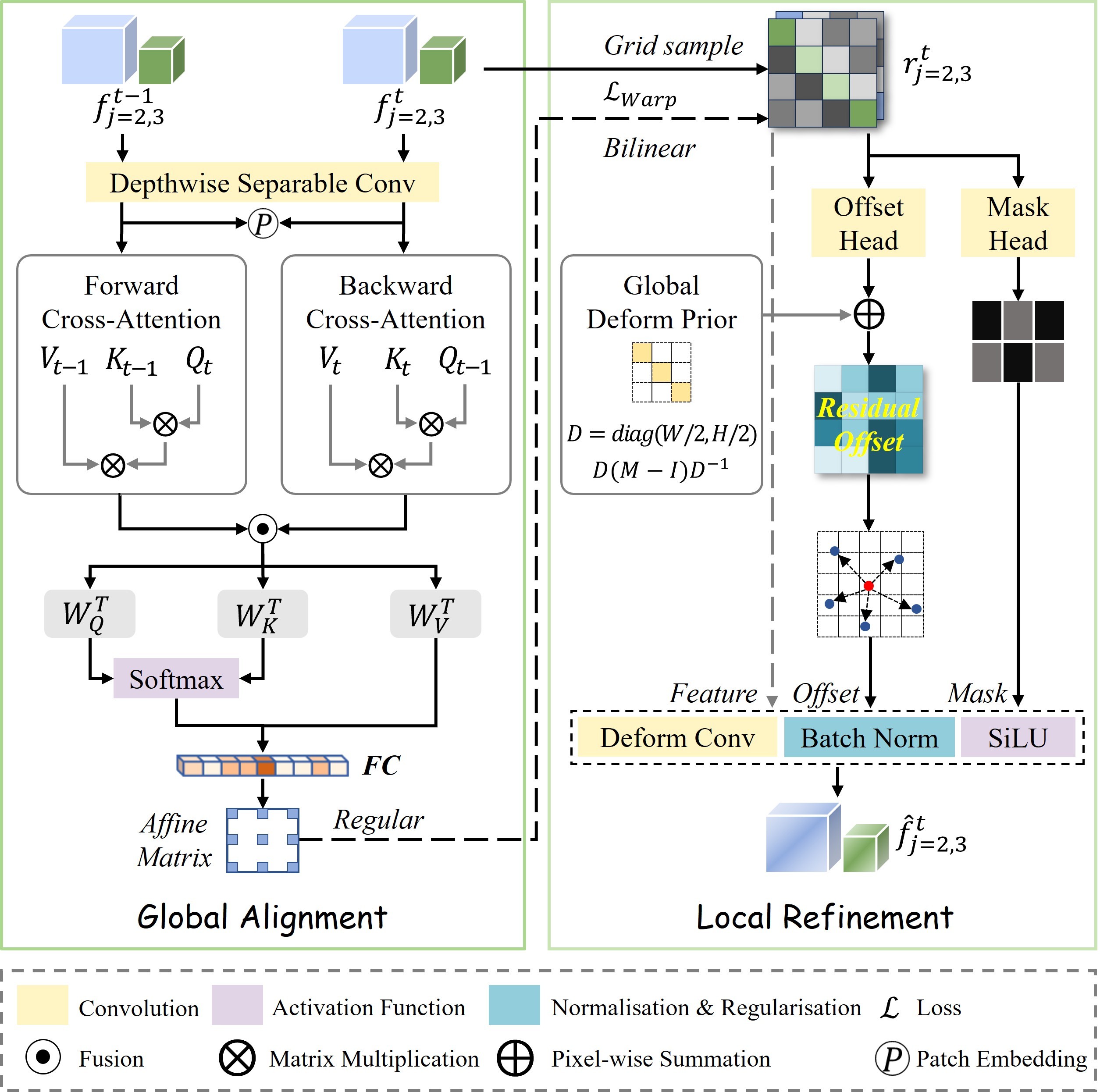}
    \caption{The structure of the GLMD module comprises a global alignment section on the left, responsible for capturing global changes between adjacent frames; and a local refinement section on the right, which processes local motion details based on the global alignment.}
    \label{fig:GLMD}
\end{figure}

As illustrated in Fig.~\ref{fig:GLMD}, the proposed GLMD module adopts a dual-branch architecture: a global alignment branch that learns affine transformations in the feature space to suppress platform-dominated background motion, and a local refinement branch that models target-intrinsic motion via adaptive, deformable sampling.

\subsubsection{Global Alignment} In the global alignment branch, features are first processed by lightweight spatial encoders built on depthwise separable convolutions to extract fundamental contextual information. The current frame and its preceding frame are then treated as query sources for forward cross-attention (FCA) and backward cross-attention (BCA), respectively. This bidirectional attention design captures temporal dependencies in both directions, enabling richer spatio-temporal feature interactions. The outputs of FCA and BCA are fused and subsequently passed to a self-attention module equipped with positional encodings to preserve spatial awareness. The self-attention mechanism computes global correlations among the queries (Q), keys (K), and values (V) across adjacent frames, thereby facilitating robust cross-frame semantic alignment under global motion. The overall multi-layer cross-attention process can be formulated as:
\begin{align}
Atten_{FCA} &= \mathrm{softmax}\left(\frac{Q_{t} K_{t-1}^{T}}{\sqrt{d}}\right) V_{t-1}, \\
Atten_{BCA} &= \mathrm{softmax}\left(\frac{Q_{t-1} K_{t}^{T}}{\sqrt{d}}\right) V_{t}, \\
Q_h, K_h, V_h &= \mathcal{M}(Atten_{FCA}, Atten_{BCA}), \\
Atten &= \mathrm{softmax}\left(\frac{Q_h K_h^T}{\sqrt{d}}\right) V_h,
\end{align}
where $ \mathcal{M}(\cdot)$ represents an MLP layer, $Q_h$, $K_h$, $V_h$ represent the projection after fusing the outputs of FCA and BCA. The tuples ($Q_t$, $K_t$, $V_t$) and ($Q_{t-1}$, $K_{t-1}$, $V_{t-1}$) correspond to the projections of the current frame and the reference frame, respectively.

Finally, the global features are projected into a low-dimensional space to regress the global affine transformation matrix $R_t\in \mathbb{R}^{2\times3}$ via a fully connected layer: 
\begin{equation}
R_t = \mathcal{R}(\mathbf[FC(\mathcal{F}(Atten)],
\end{equation}
where $\mathcal{R}(\cdot)$ is a reshape operation, $FC(\cdot)$ denotes a fully connected layer, and $\mathcal{F}(\cdot)$ denotes dimensionality reduction.

\subsubsection{Local Refinement}Guided by the transformation prior learned in the global alignment branch, the local refinement branch further employs deformable convolutions and masking mechanisms to compensate for residual local motion and achieve fine-grained spatial alignment, thereby improving adaptability to unstabilized videos and improves the recovery of object details. Specifically, the mid-level features of the current $t$-th frame, $\{f^t_j\mid j=2,3\}$, are first coarsely aligned via the transformation matrix $R_t$ estimated by the global branch, yielding the globally compensated features $\{r_j^t \mid j=2,3\}$ containing global motion information. These features are then fed into the deformable convolution for local refinement. The global warping operation is formulated as:
\begin{equation}
r_j^t = Warp(R_t,f_j^t),
\end{equation}
where $Warp(\cdot)$ denotes bilinear interpolation with grid sampling.

Next, the global transformation matrix $R_t$ is decomposed into a linear component $A\in \mathbb{R}^{2\times2}$ and a translation vector $b \in \mathbb{R}^2 $. To incorporate the affine alignment results into local alignment, we construct a fundamental offset term $\Delta{P_k}$ for each convolution kernel position $P_k$ in feature space. To accommodate variations in feature resolution, we perform spatial scale normalization using a diagonal matrix: $D= diag(W/2,H/2)$, where $W$ and $H$ represent the image width and height, respectively. The resulting affine prior offset $\Delta{P_k}$ provides a principled initialization for deformable sampling, upon which the local branch learns fine-grained corrections:
\begin{equation}
\Delta{P_k} = D(A-I)D^{-1}P_k.
\end{equation}

Additionally, a convolutional neural network is employed to predict a data-driven residual offset $\Delta{P_k^{res}}$ to compensate for local motion residuals:
\begin{equation}
\Delta{P_k^{res}},m = Mask(r_j^t),
\end{equation}
where $Mask(\cdot)$ denotes a $3\times3$ modulation mask head, and $m$ is the modulation mask activated by a sigmoid function.

Finally, the motion-aware feature map $r^t_j$, the total offset $\mathcal{O} $, and the modulation mask $m$ are fed into a deformable convolution layer. The resulting features are normalized by batch normalization to obtain the refined feature map $\{\hat{f}_j^t \mid j=2,3\}$, which incorporates both global alignment and local refinement. This process is expressed as:
\begin{align}
\mathcal{O} &= \Delta{P_k}+\Delta{P_k^{res}},\\
\hat{f}_j^t &= SL(BN(De(r^t_j,\mathcal{O} ,m))),
\end{align}
where $BN(\cdot)$, $SL(\cdot)$, and $De(\cdot)$ denote batch normalization, sigmoid linear unit, and deformable convolution, respectively.

\subsection{Temporal Dependency Feature Pyramid}

Temporal dependency modeling is crucial for tiny-object tracking in satellite videos, as aggregating motion cues across consecutive frames mitigates noise, weak appearance features, and occlusions, especially under limited spatial resolution and subtle target motion.
Among existing temporal modeling strategies, ConvGRU~\cite{ConvGRU_2015} provides a superior trade-off between robustness and computational efficiency compared to optical-flow-based and Transformer-based methods.
When inter-frame displacement is smaller than two pixels and target motion is subtle, optical flow estimation often becomes unreliable. Moreover, optical-flow-based methods typically rely on explicit motion estimation via feature matching or cost volume construction, which introduces additional computational overhead. In contrast, although Transformer-based models are effective in capturing long-range dependencies, they usually incur high computational complexity when applied to video sequences.
Therefore, we adopt ConvGRU to efficiently model temporal dependencies.

As illustrated in the lower part of Fig.~\ref{fig:network}, the feature sequence $\{\hat{f}_i^{\tau} = (f_1^{\tau}, \hat{f}_2^{\tau}, \hat{f}_3^{\tau}, f_4^{\tau}) \mid \tau = t-k+1, \dots, t \}$ is first processed by ConvGRU to obtain a temporally fused feature set $\{\tilde{f}_{i}^{\tau}\}$. Given the input $\{\hat{f}_{i}^t \mid i = 1,2,3,4\}$ in the sequence and the hidden state $\{\hat{f}_{i}^{t-1} \mid i = 1,2,3,4\}$  from the previous frame, ConvGRU computes the update and reset gates followed by the candidate state and the new hidden state:
\begin{equation}
\begin{aligned}
    z &= \sigma{[W_z(\hat{f}_i^{\tau},h)]},\\
    r &= \sigma{[W_r(\hat{f}_i^{\tau},h]},\\
    \tilde{h} &=tanh\{ \sigma{[W_h(\hat{f}_i^{\tau},r\odot{h_{}})]}\},\\
    h &=(1-z)\odot{h_{}}+z\odot{\tilde{h}},
\end{aligned}
\end{equation}
where $z$ and $r$ denote the update gate and reset gate, respectively, $W_z$, $W_r$, and $W_h$ correspond respectively to the linear transformation weight matrices for different gates, $\sigma(\cdot)$ is the Sigmoid function, $tanh$ is the tangent function and $\odot$ indicates the element-wise multiplication.

Next, to further enhance the representation of tiny objects, the multi-level temporal feature maps $\{\tilde{f}_{i}^{\tau} \mid  \tau=t-k+1,\ldots,t; i = 1,2,3,4\}$ are fed into a bidirectional feature pyramid network (BiFPN)~\cite{BiFPN_2020}, yielding a fused multi-scale feature set $\{F^{\tau}\mid \tau=t-k+1,\ldots,t\}$.
Through weighted bidirectional information flow, BiFPN adaptively balances feature contributions from different scales, preserving high-resolution spatial details that are critical for tiny-object localization while maintaining strong semantic consistency from deeper layers.

Finally, to capture the dynamic evolution of objects in a higher-level semantic space, we apply a second ConvGRU aggregation over the fused feature sequence $\{F^{\tau} \mid \tau=t-k+1,\ldots,t\}$.
This temporal aggregation establishes a cross-frame memory mechanism within the feature space, thereby implicitly reinforcing identity consistency modeling. It not only further stabilizes object representations but also significantly enhances the robustness of cross-frame associations. Subsequently, these features are utilized to generate the heatmap, width-height regression, and offset regression branches, enabling joint object detection and tracking.

\subsection{Objective Functions}
To enhance cross-frame feature consistency, we introduce a warp alignment loss that is applied after affine-based global alignment. This loss explicitly enforces semantic consistency between the aligned features of the query frame and those of the reference frame in the feature space, thereby stabilizing feature representations across frames. Let $f_{re}$ denote the feature map of the reference frame, and $f_{warp}$ be the feature map of the query frame after affine warping. The warp alignment loss is defined as:
\begin{equation}
\mathcal{L}_{warp} = \mathbb{E}_{p \in \Omega} \left[ 1 - \frac{\langle f_{warp}(p), f_{re}(p) \rangle}{\lVert f_{warp}(p) \rVert_2 \lVert f_{re}(p) \rVert_2 + \epsilon} \right],
\end{equation}
where $\langle\cdot\rangle$ denotes the inner product, $\Omega$ represents the spatial domain of the feature map, and $\epsilon$ is a small constant for numerical stability. This cosine similarity-based formulation encourages aligned features to remain semantically consistent while being invariant to feature magnitude.

Furthermore, we adopt a combination of focal loss $\mathcal{L}_{focal}$ for classification~\cite{Focal_loss} and regression loss $\mathcal{L}_1$ for bounding-box prediction~\cite{L1_loss}. The overall training objective is obtained by jointly optimizing the proposed warp alignment loss together with the two detection losses:
\begin{equation}
    \mathcal{L}=\mathcal{L}_{warp}+\alpha{\mathcal{L}_{focal}}+\beta{\mathcal{L}_1},
\end{equation}
where $\alpha$ and $\beta$ are weighting factors.

\section{Experiment Results}
In this section, we conduct a comprehensive evaluation of the proposed DeTracker. We first describe the experimental setup, followed by quantitative and qualitative comparisons with state-of-the-art methods. Finally, we conduct ablation studies to assess the contribution of each key component.

\subsection{Experimental Setup}
\subsubsection{Datasets}
The proposed method is evaluated on two types of data: (i) simulated unstabilized dataset \emph{SDM-Car-SU}, which covers diverse background motion patterns with varying speeds (1–3 pixels per frame) and motion directions, enabling systematic assessment under different levels of unstabilized motion; and (ii) real unstabilized satellite videos acquired by the Luojia-3 01 satellite to validate its effectiveness in practical scenarios.

\subsubsection{Baselines} 

We compare the proposed DeTracker with a range of state-of-the-art trackers, including both TBD-based approaches, such as DeepSORT~\cite{Wojke2017Simple}, CKDNet-SMTNet~\cite{FENG2021Cross}, and ByteTrack~\cite{Bytetrack}, and JDT-based approaches, including CenterTrack~\cite{Zhou2020Tracking}, FairMOT~\cite{Zhang2021FairMOT}, DSFNet~\cite{Xiao2022DSFNet}, MP2Net~\cite{Zhao2024MP2Net}, DAF~\cite{Yu2025On-Satellite}, PIFTrack~\cite{Chen2025PIFTrack}, and MOSAIC-Tracker~\cite{Zou2025MOSAIC}. 

To ensure fair comparisons, all methods are trained and evaluated under the same experimental settings, with hyperparameters and training configurations adopted from the original papers or official open-source implementations.

\subsubsection{Evaluation Metrics}
For satellite video multi-object tracking, we adopt MOTA to measure overall tracking performance, particularly reflecting detection accuracy, and IDF1 to evaluate identity consistency, which reflects the quality of track association:
\begin{align}
\mathrm{MOTA} &= 1- \frac{FN+FP+ID_S}{GT},\\
\mathrm{IDF1}&=\frac{2 \times ID_{TP}}{2 \times ID_{TP}+ID_{FP}+ID_{FN}},
\end{align}
where $FP$, $FN$, and $ID_S$ denote false positives, false negatives, and identity switches, respectively; $GT$ is the number of ground-truth objects; $ID_{TP}$, $ID_{FP}$, and $ID_{FN}$ are the numbers of true-positive, false-positive, and false-negative identity matches.
In addition, we report identity precision and recall (IDP and IDR), as well as standard MOT statistics including Precision (Prcn), Recall (Rcll), Mostly Tracked (MT), and Mostly Lost (ML), which provide complementary insight into tracking quality and robustness.

\begin{table*}[t]
\centering
\caption{Quantitative comparisons on \emph{SDM-Car-SU} Dataset (U1–U3 correspond to increasing platform motion intensities). The best and the second-best results are in \textbf{\textcolor{red}{bold red}} and \textbf{\textcolor{blue}{bold blue}}, respectively. $\uparrow$ indicates higher is better, while $\downarrow$ indicates lower is better.}
\label{tab:Quantitative}
\renewcommand{\arraystretch}{1.2}
\setlength{\tabcolsep}{4pt}
\begin{tabular}{l| c c c c c | c c c c c| c c c c c}
\toprule
\multirowcell{2}{{Method}} & 
\multicolumn{5}{c|}{\textbf{U1}} & 
\multicolumn{5}{c|}{\textbf{U2}} & 
\multicolumn{5}{c}{\textbf{U3}} \\
\cmidrule(lr){2-6} \cmidrule(lr){7-11} \cmidrule(lr){12-16}
& MOTA$\uparrow$ & IDF1$\uparrow$ & IDs$\downarrow$ & FP$\downarrow$ & FN$\downarrow$ 
  & MOTA$\uparrow$ & IDF1$\uparrow$ & IDs$\downarrow$ & FP$\downarrow$ & FN$\downarrow$ 
  & MOTA$\uparrow$ & IDF1$\uparrow$ & IDs$\downarrow$ & FP$\downarrow$ & FN$\downarrow$ \\
\midrule
DeepSORT~\cite{Wojke2017Simple}  & 39.7\% & 55.8\% & \textbf{\textcolor{red}{1720}} & 54105 & 35205 & 42.2\% & 62.8\% & 2829 & 41045 & 56879 & 14.4\% & 39.0\% & 6154 & 81293 & 54593 \\

CenterTrack~\cite{Zhou2020Tracking}  & 40.4\% & 62.7\% & 3690 & 41906 & 38117 & 37.5\% & 61.6\% & 3045 & 46028 & 57090 & 25.3\% & 53.7\% & 5850 & 75863 & 48243 \\

FairMOT~\cite{Zhang2021FairMOT}  & 36.1\% & 60.8\%  & 4013 & 79904 & 49417 & 31.5\% & 60.2\% & 2942 & 68064 & 54656 & 21.7\% & 50.3\% & 5484 & 70121 & 53650 \\

CKDNet-SMTNet~\cite{FENG2021Cross}  & 43.5\% & 56.7\% & 3294 & 54283 & 37757 & 42.8\% & 62.6\%  &\textbf{\textcolor{blue}{2754}}  & 44593 & 59550 & 40.3\% & 55.1\% &\textbf{\textcolor{blue}{4117}} & 39904 & \textbf{\textcolor{blue}{46773}}  \\

ByteTrack~\cite{Bytetrack}  & 42.7\% & 57.6\% & 3797 & 56833 & 37251 & 44.6\% & 63.9\% & 2894 & 39472 & 56620 & 35.7\% & 55.6\% & 4924 & 50538 & 56375 \\

DSFNet~\cite{Xiao2022DSFNet}  & 46.2\% & 63.8\% & 3614 & 57304 & 34487 & 42.9\% & 62.2\% & 2976 & 38911 & 55761 & 32.6\% & 54.3\% & 5476 & 58711 & 57709 \\

MP2Net~\cite{Zhao2024MP2Net}  & 51.7\% & 66.8\% & 2105 & 50939 &\textbf{\textcolor{blue}{33274}} & 49.2\% & 64.5\% & 2925 & 35616 & 56204 & 42.0\% & 57.2\% & 4214 & 31399 & 55254 \\

DAF~\cite{Yu2025On-Satellite}  & 51.3\% & 66.5\% & 3230 & 51900 & 34996 & 48.3\% & 64.9\% & 2853 & 36021 & 56957 & \textbf{\textcolor{blue}{45.5\%}} & \textbf{\textcolor{blue}{61.3\%}}  & 4171 & 38584 & 48751 \\

PIFTracker~\cite{Chen2025PIFTrack}  &\textbf{\textcolor{blue}{58.8\%}}  & \textbf{\textcolor{blue}{68.4\%}}  & 2205 & \textbf{\textcolor{red}{22949}} & 36815 & \textbf{\textcolor{blue}{53.7\%}}   &\textbf{\textcolor{blue}{66.5\%}}  & 2760 & \textbf{\textcolor{red}{31893}} & 58854 & 41.1\% & 56.2\% & 4206 & \textbf{\textcolor{blue}{31123}}& 47092 \\

MOSAIC-Tracker~\cite{Zou2025MOSAIC}  & 53.1\% & 66.5\% & 3106 & 38997 & 42687 & 48.0\% & 65.9\% & 2918 & 43041 & \textbf{\textcolor{blue}{54338}}& 32.7\% & 54.1\% & 5058 & 53607 & 58506 \\

\midrule
\textbf{DeTracker (Ours)}  
& \textbf{\textcolor{red}{61.1\%}} & \textbf{\textcolor{red}{68.7\%}} &\textbf{\textcolor{blue}{1863}} &\textbf{\textcolor{blue}{23065}} & \textbf{\textcolor{red}{33046}} 
& \textbf{\textcolor{red}{55.3\%}} & \textbf{\textcolor{red}{67.2\%}} & \textbf{\textcolor{red}{2657}} & \textbf{\textcolor{blue}{34330}}& \textbf{\textcolor{red}{51570}}
& \textbf{\textcolor{red}{52.4\%}} & \textbf{\textcolor{red}{63.8\%}} & \textbf{\textcolor{red}{3231}} & \textbf{\textcolor{red}{14821}} & \textbf{\textcolor{red}{19775}} \\
\bottomrule
\end{tabular}
\end{table*}

\begin{table*}[!htbp]
  \centering
  \caption{Quantitative results on real Unstabilized satellite videos. The best and the second-best results are indicated in \textbf{\textcolor{red}{bold red}} and \textbf{\textcolor{blue}{bold blue}}, respectively. $\uparrow$ indicates higher is better, while $\downarrow$ indicates lower is better.}
  \label{tab:lj}
  \begin{tabular*}{\textwidth}{@{\extracolsep{\fill}}lccccccccccc}
    \toprule
    Method & MOTA$\uparrow$ & IDF1$\uparrow$ & IDP$\uparrow$ & IDR$\uparrow$ & Rcll$\uparrow$ & Prcn$\uparrow$ & IDs$\downarrow$ & FP$\downarrow$ & FN$\downarrow$& MT$\uparrow$ & ML$\downarrow$\\
    \midrule
     DSFNet~\cite{Xiao2022DSFNet} & 29.9\% & 56.6\% & 59.7\% & 53.8\% & 60.7\% & 67.3\% & 263 & 6126 & 8158 & \textbf{\textcolor{blue}{133}} & 107 \\
     
     MP2Net~\cite{Zhao2024MP2Net} & 36.4\% & \textbf{\textcolor{blue}{61.7\%}} & 65.2\% & \textbf{\textcolor{blue}{58.5\%}}& \textbf{\textcolor{blue}{63.7\%}} & 70.9\% & 357 & 7270 & 10123 & \textbf{\textcolor{red}{178}} & 75 \\ 
     
     DAF~\cite{Yu2025On-Satellite} & 31.9\% & 53.8\% & 59.9\% & 48.8\% & 57.4\% & 70.4\% & \textbf{\textcolor{blue}{154}} & 4443 &\textbf{\textcolor{blue}{5304}} & 95 & \textbf{\textcolor{blue}{40}} \\  
     
     PIFTrack~\cite{Chen2025PIFTrack} & \textbf{\textcolor{blue}{37.0\%}} & 60.1\% & \textbf{\textcolor{red}{71.8\%}} & 53.0\% & 58.4\% & \textbf{\textcolor{red}{79.1\%}} & 192 & \textbf{\textcolor{blue}{4282}} & 6171 & 76 & 42 \\ 
     
     MOSAIC-Tracker~\cite{Zou2025MOSAIC} & 35.2\% & 57.2\% & 64.7\% & 51.3\% & 57.8\% & 73.0\% & 266 & 4431 & 8749 & 121 & 111 \\  
     
     \midrule
     \textbf{DeTracker (Ours)}  & \textbf{\textcolor{red}{45.3\%}} & \textbf{\textcolor{red}{67.8\%}} & \textbf{\textcolor{blue}{69.3\%}}& \textbf{\textcolor{red}{70.25\%}} & \textbf{\textcolor{red}{74.8\%}} & \textbf{\textcolor{blue}{73.9\%}} & \textbf{\textcolor{red}{182}} & \textbf{\textcolor{red}{4109}} & \textbf{\textcolor{red}{3909}} & 127 & \textbf{\textcolor{red}{22}} \\   
    \bottomrule
  \end{tabular*}
\end{table*}

\subsubsection{Implementation Details}
Prior to training, standard data augmentation techniques were applied, including inversion, random flipping, and rotation. The initial learning rate was set to $1.25\times10^{-4}$ and decayed using a cosine annealing schedule. The model was trained for 150 epochs with a batch size of 3. The loss weighting parameters $\alpha$ and $\beta$ were set to 0.2 and 0.3, respectively. The number of input frames $k$ was set to 5. All experiments were conducted on a single NVIDIA A100 Tensor Core GPU.

\subsection{Comparison with the State-of-the-Art}

\subsubsection{Quantitative Analysis}
Table~\ref{tab:Quantitative} presents the quantitative comparison on the \emph{SDM-Car-SU} dataset. DeTracker consistently achieves the best overall performance across all motion intensity subsets (U1--U3), where U1--U3 correspond to increasing platform motion intensities from 1 to 3 pixels per frame. 
Specifically, in terms of the two core metrics, DeTracker achieves 61.1\%, 55.3\%, 52.4\% in MOTA and 68.7\%, 67.2\%, 63.8\% in IDF1 on U1, U2, and U3, respectively, outperforming all competing methods by a clear margin.
Compared with the second-best method, PIFTrack~\cite{Chen2025PIFTrack}, DeTracker yields an average improvement of 5.1\% in MOTA and 2.9\% in IDF1 across the three motion intensity levels.
Additionally, it substantially reduces IDs, FP, and FN, indicating stronger robustness in data association and improved target recall in unstabilized videos.

To further evaluate generalization under real on-orbit conditions, we conduct experiments on real unstabilized satellite videos. 
As reported in Table~\ref{tab:lj}, DeTracker achieves 45.3\% MOTA and 67.8\% IDF1, surpassing PIFTrack by 8.3\% and 7.7\%, respectively. 
These results demonstrate that DeTracker not only performs well on simulated data but also maintains strong accuracy and robustness in real unstabilized satellite scenarios.

\tabcolsep=1pt
\begin{figure*}[!htb]
    \centering
\small{
        \begin{tabular}{ccccccc}

\includegraphics[width=0.19\textwidth,height=2.94cm,keepaspectratio]{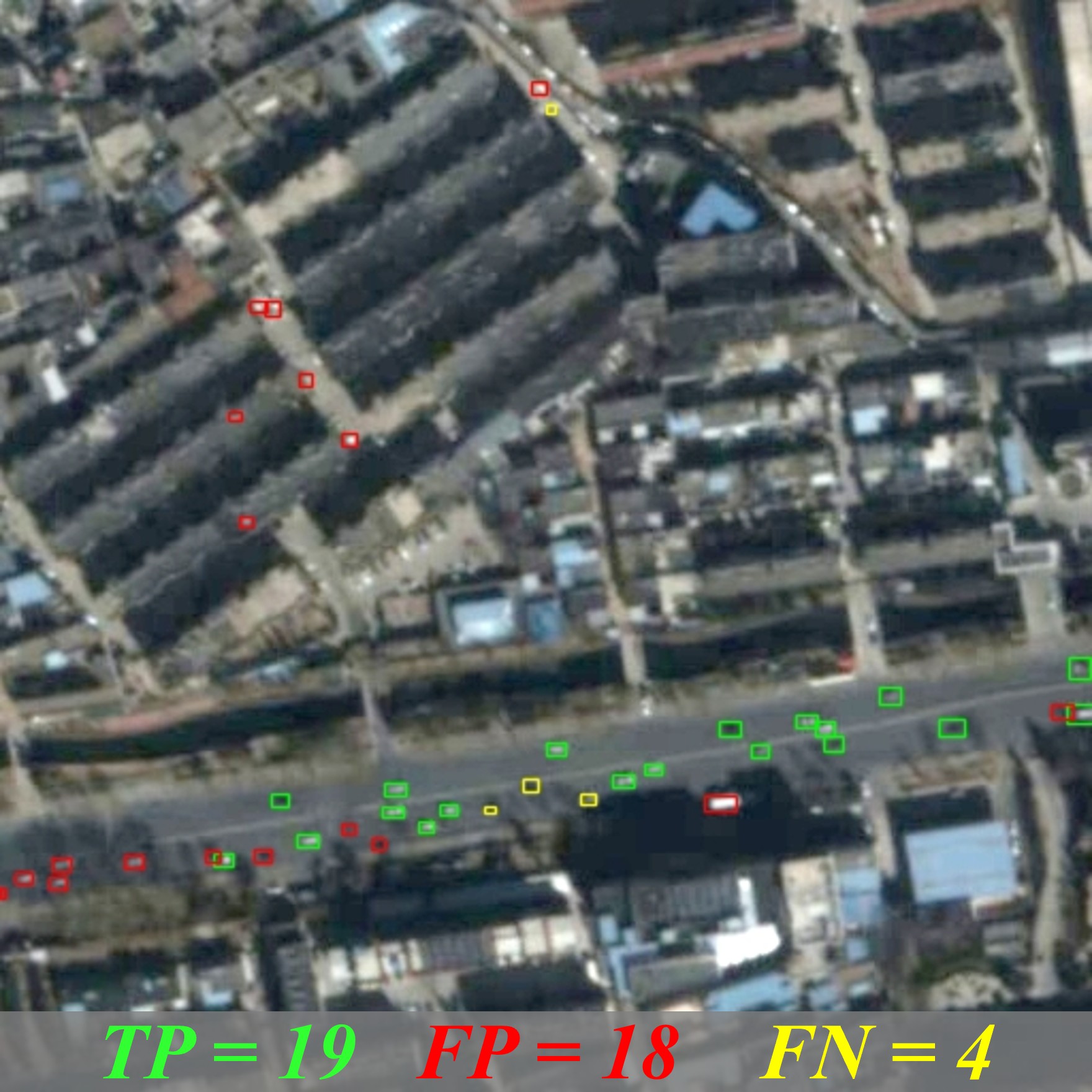} &
\includegraphics[width=0.19\textwidth,height=2.94cm,keepaspectratio]{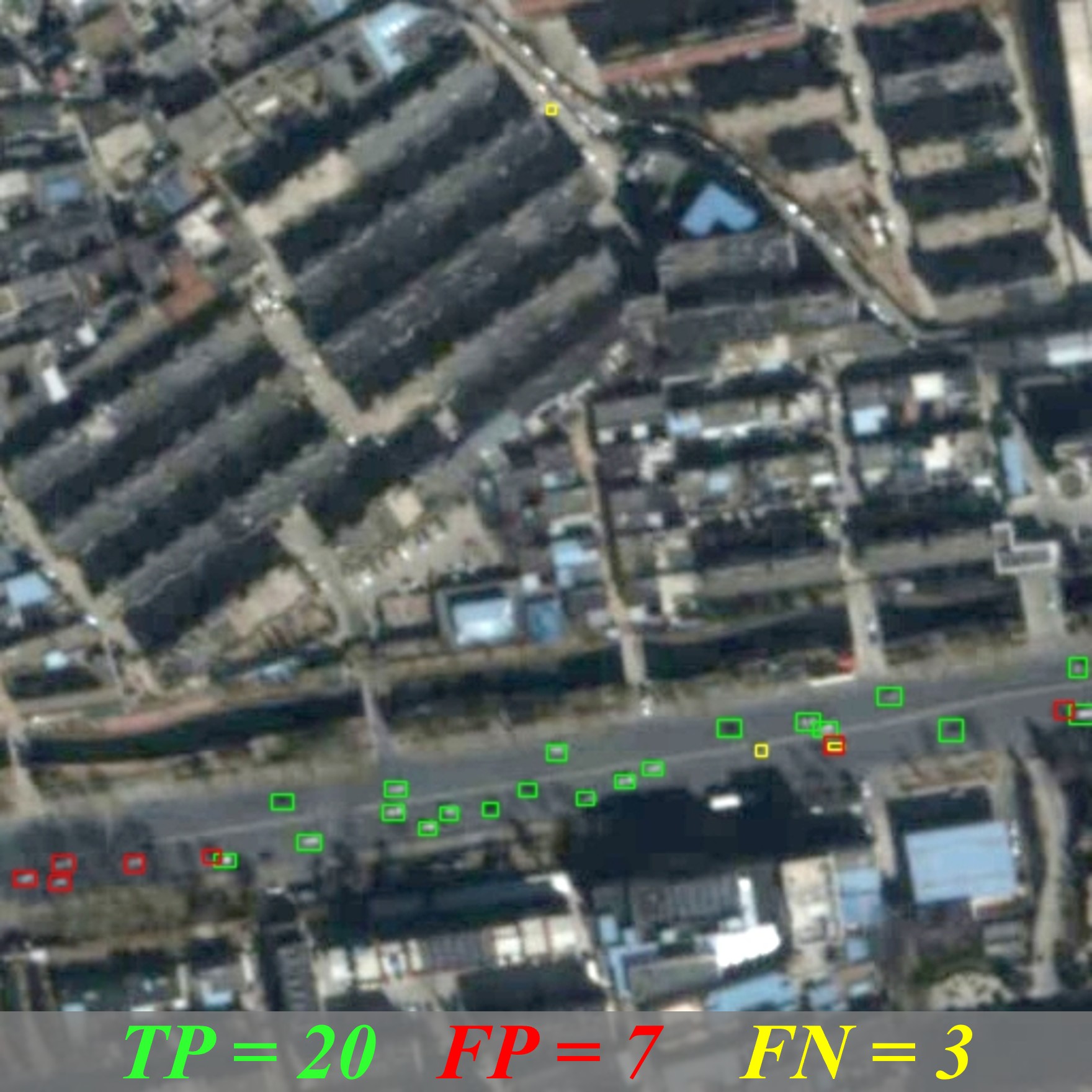} &
\includegraphics[width=0.19\textwidth,height=2.94cm,keepaspectratio]{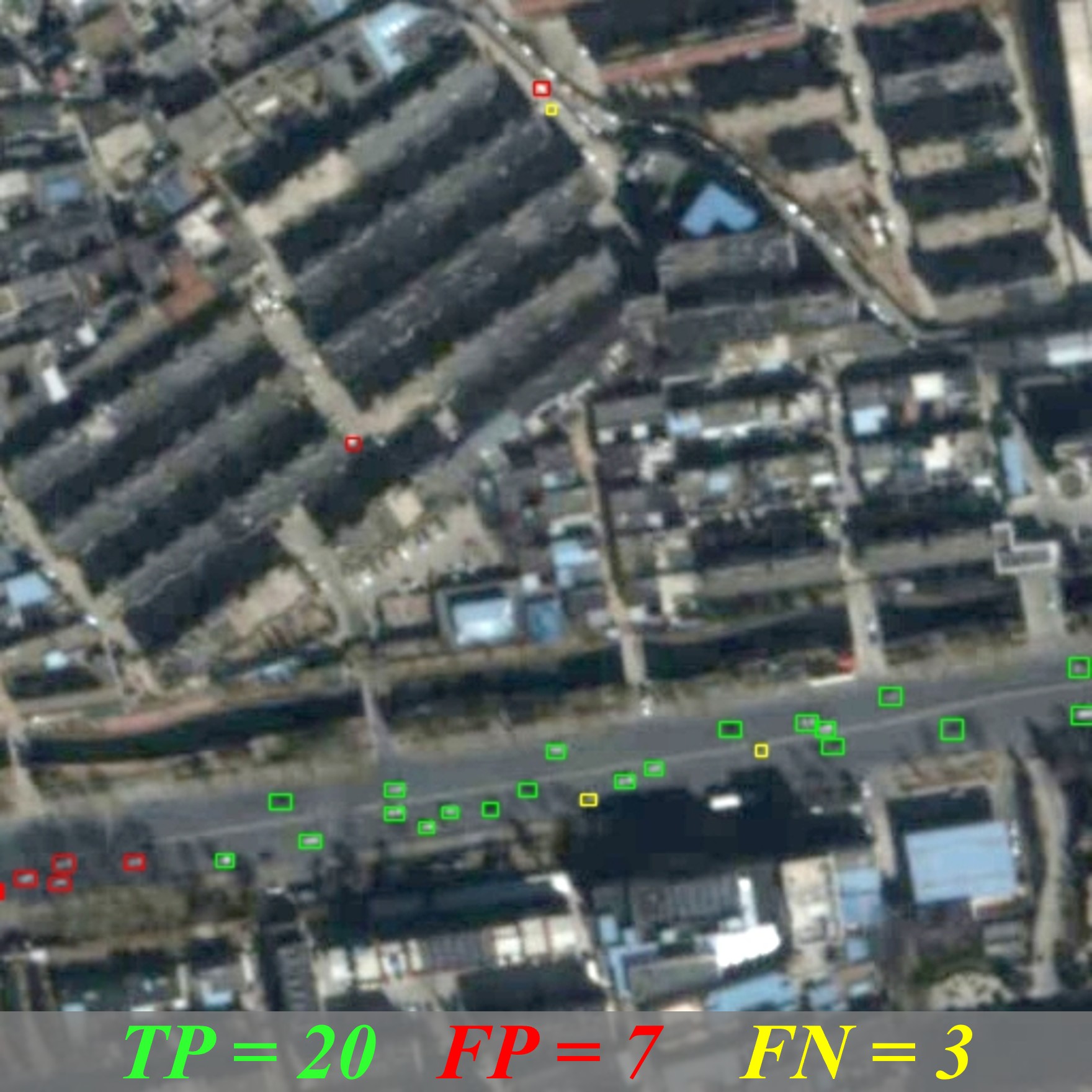} &
\includegraphics[width=0.19\textwidth,height=2.94cm,keepaspectratio]{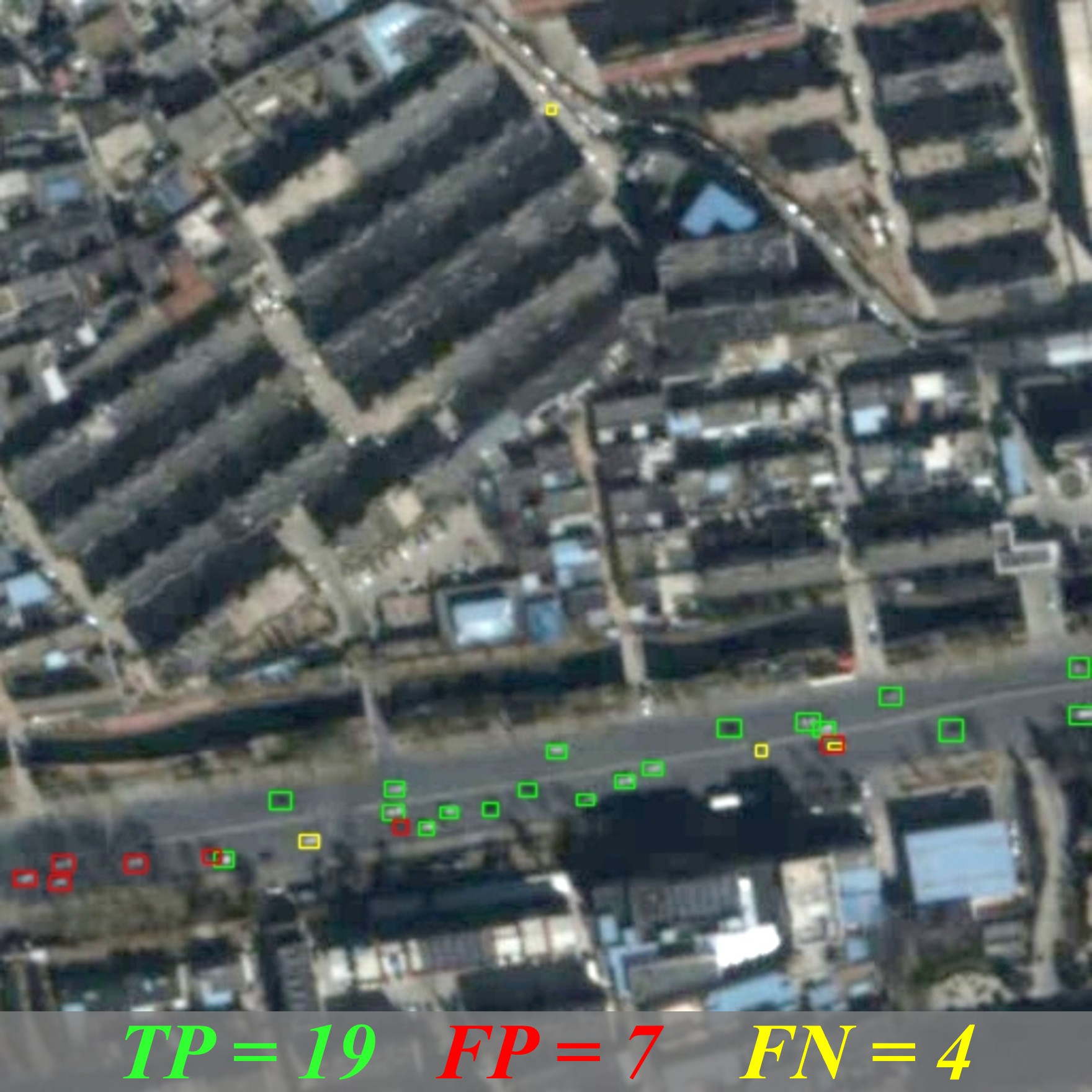} &
\includegraphics[width=0.19\textwidth,height=2.94cm,keepaspectratio]{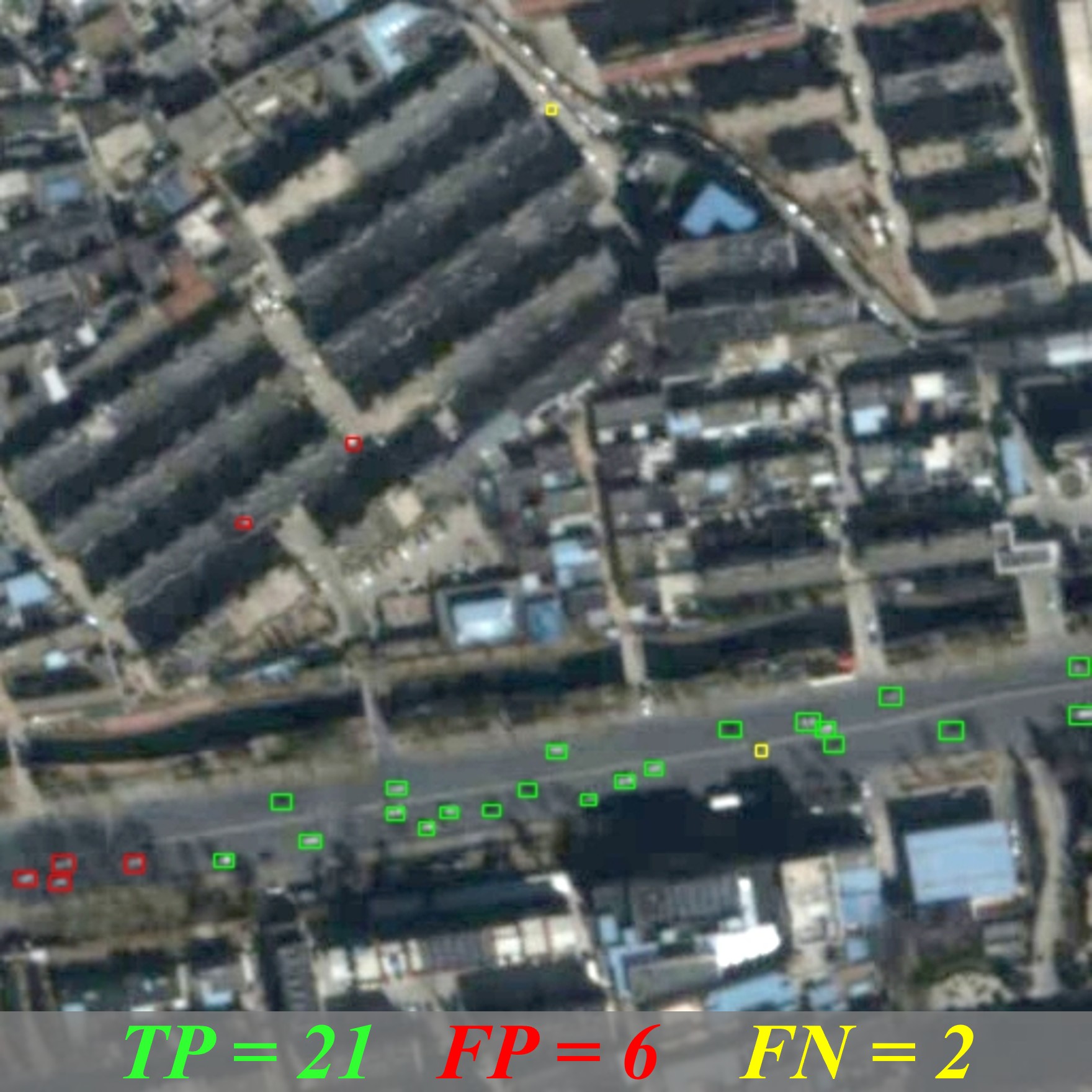} &
\includegraphics[width=0.19\textwidth,height=2.94cm,keepaspectratio]{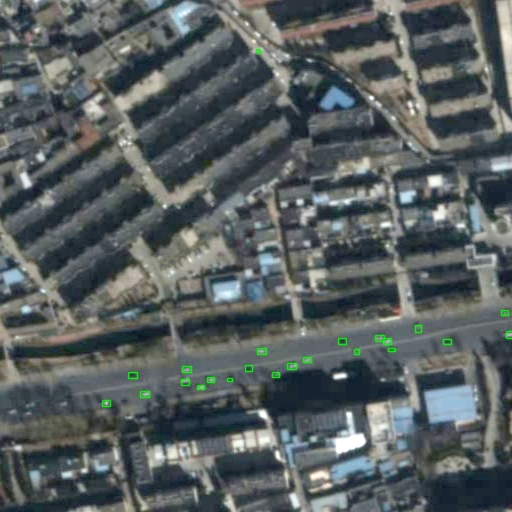} & \\
            
\includegraphics[width=0.19\textwidth,height=2.94cm,keepaspectratio]{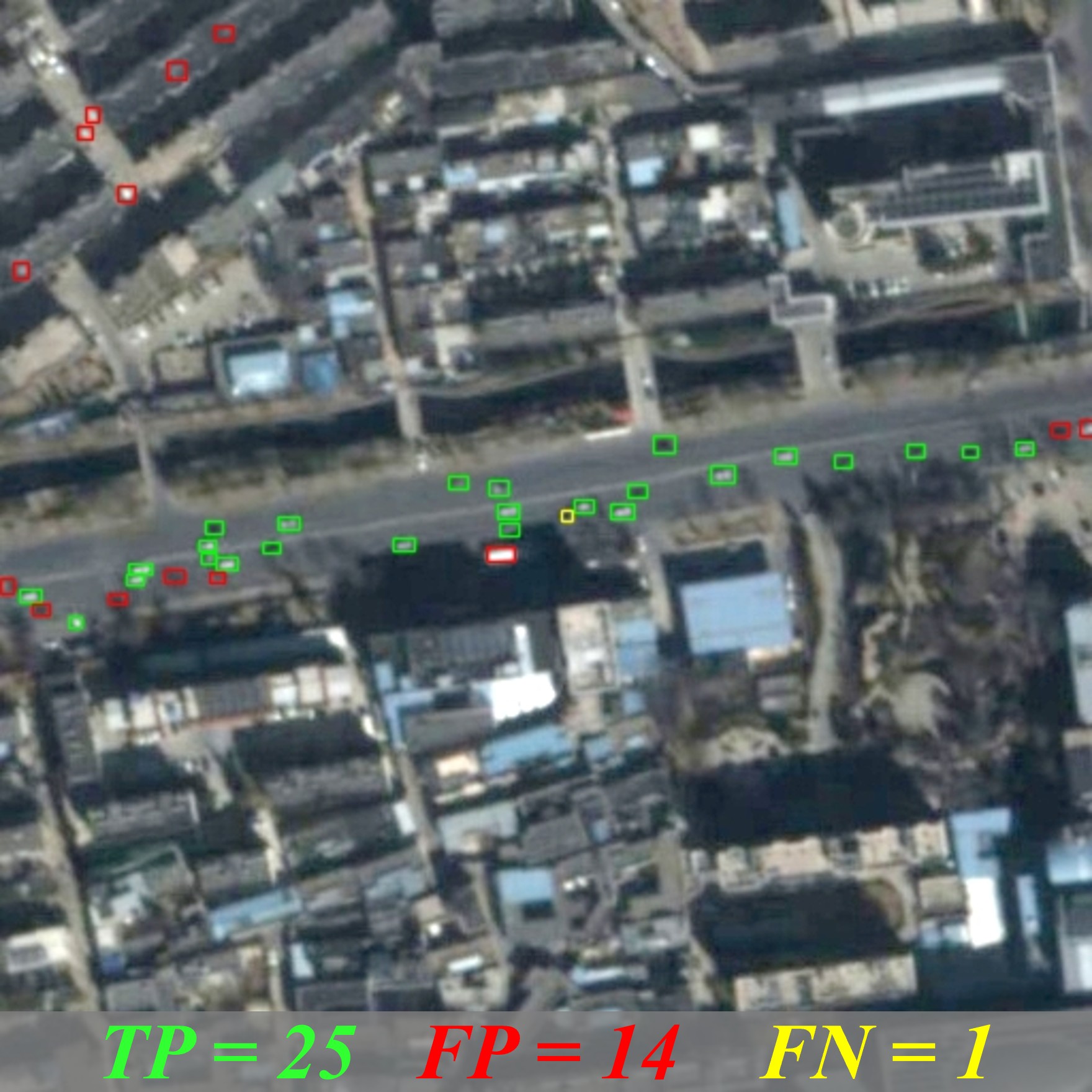} &
\includegraphics[width=0.19\textwidth,height=2.94cm,keepaspectratio]{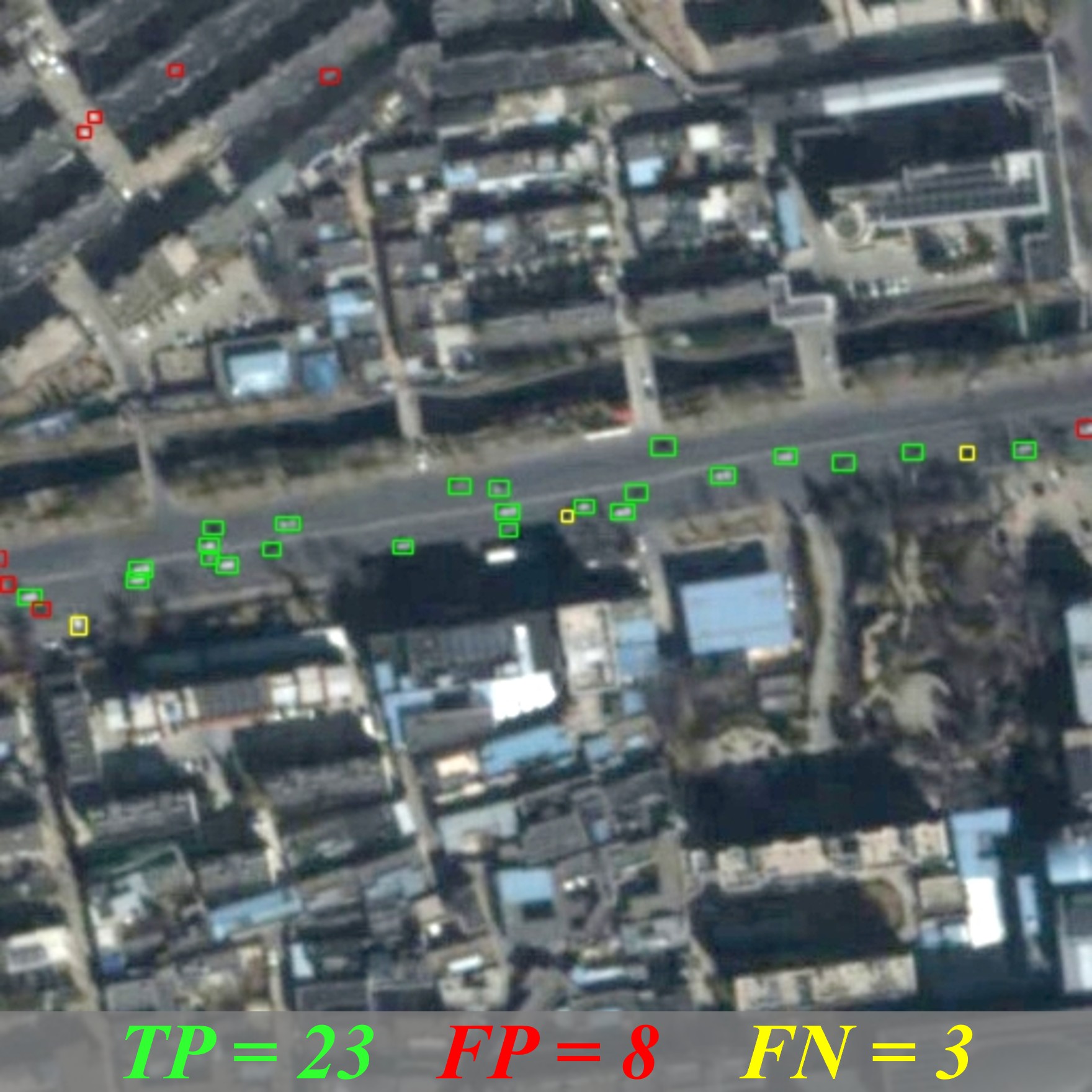} &
\includegraphics[width=0.19\textwidth,height=2.94cm,keepaspectratio]{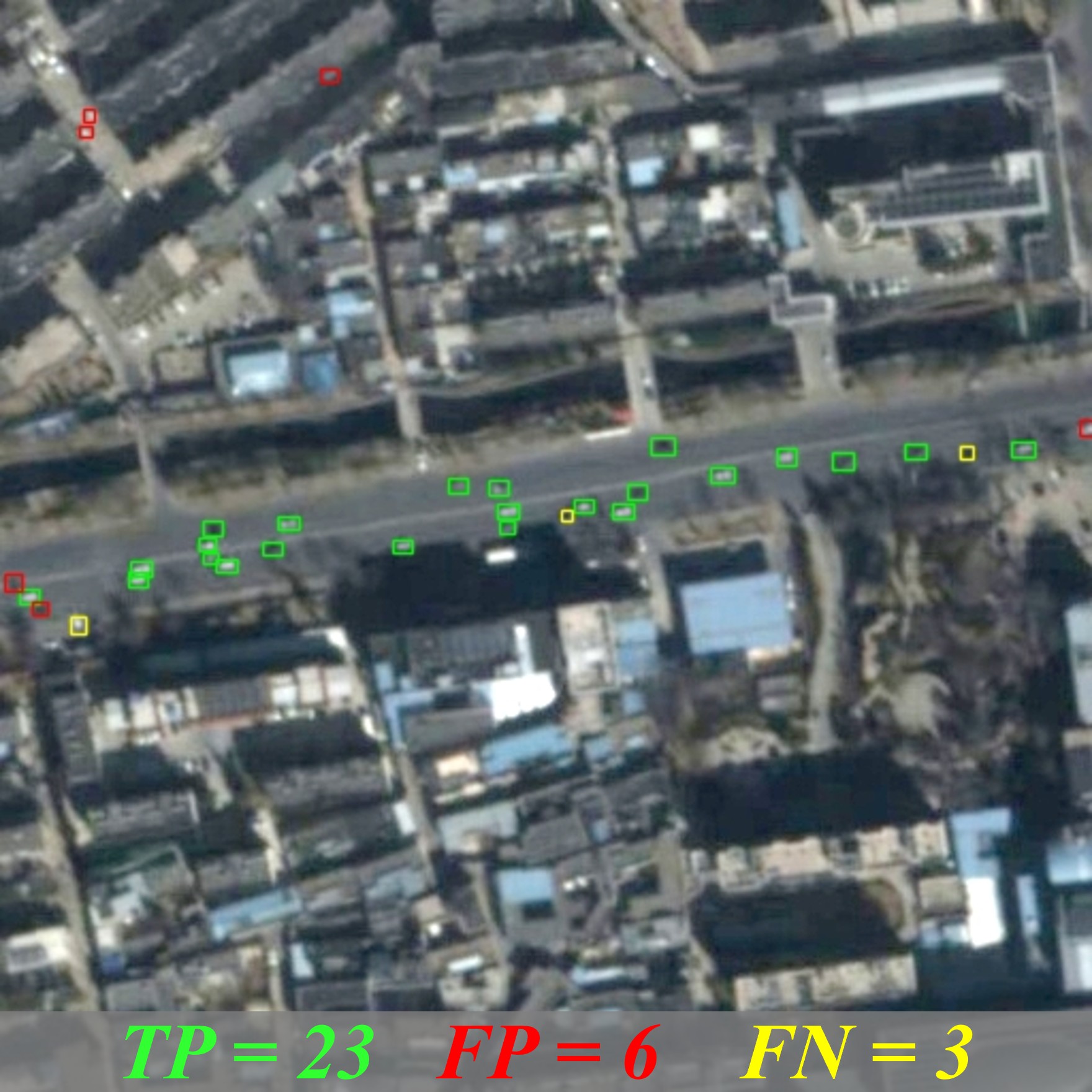} &
\includegraphics[width=0.19\textwidth,height=2.94cm,keepaspectratio]{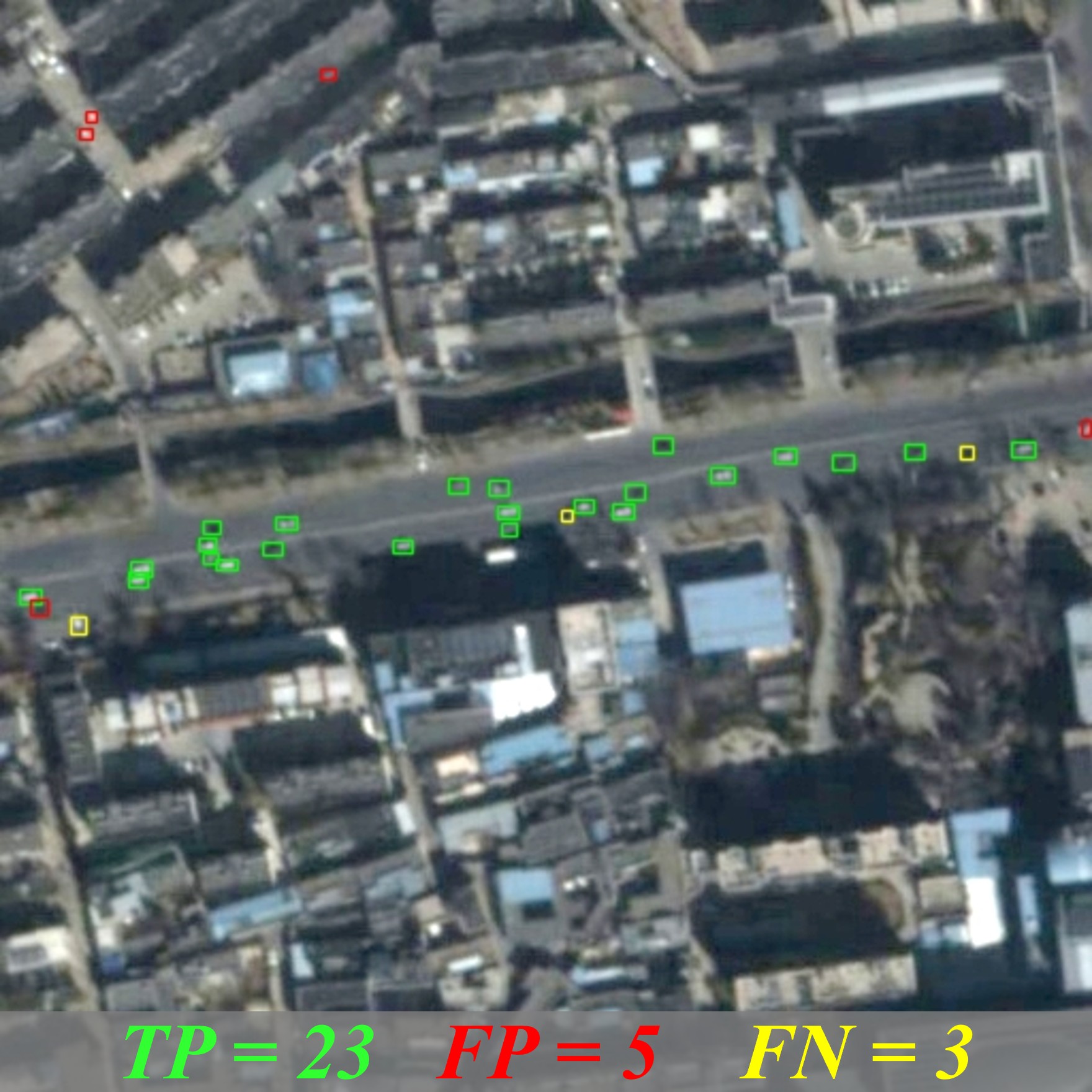} &
\includegraphics[width=0.19\textwidth,height=2.94cm,keepaspectratio]{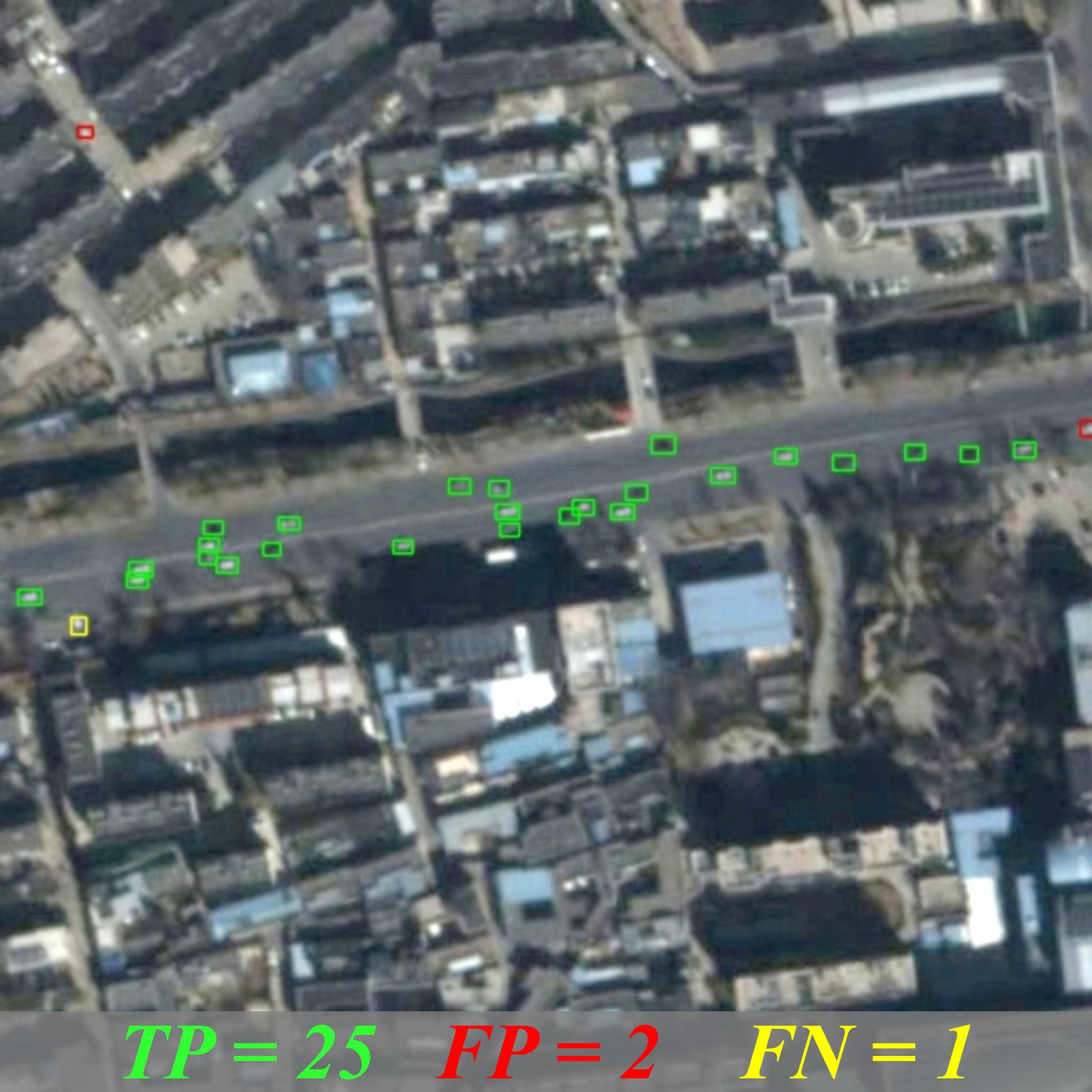} &
\includegraphics[width=0.19\textwidth,height=2.94cm,keepaspectratio]{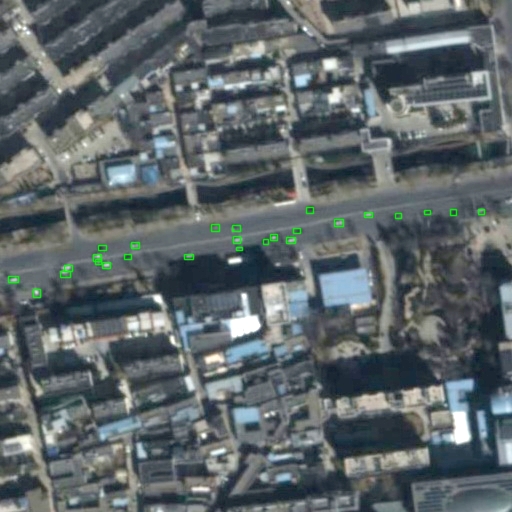} & \\
          
\includegraphics[width=0.19\textwidth,height=2.94cm,keepaspectratio]{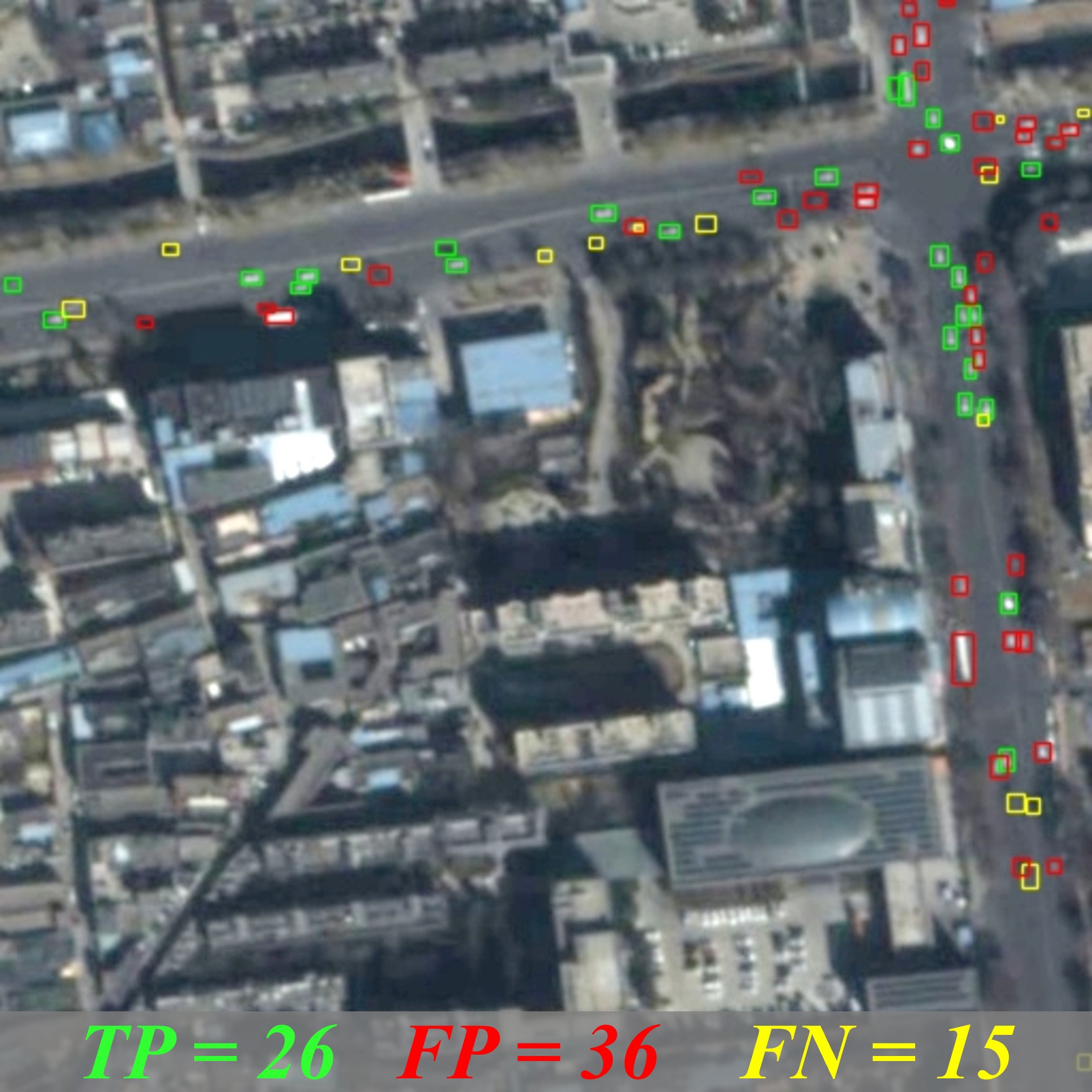} &
\includegraphics[width=0.19\textwidth,height=2.94cm,keepaspectratio]{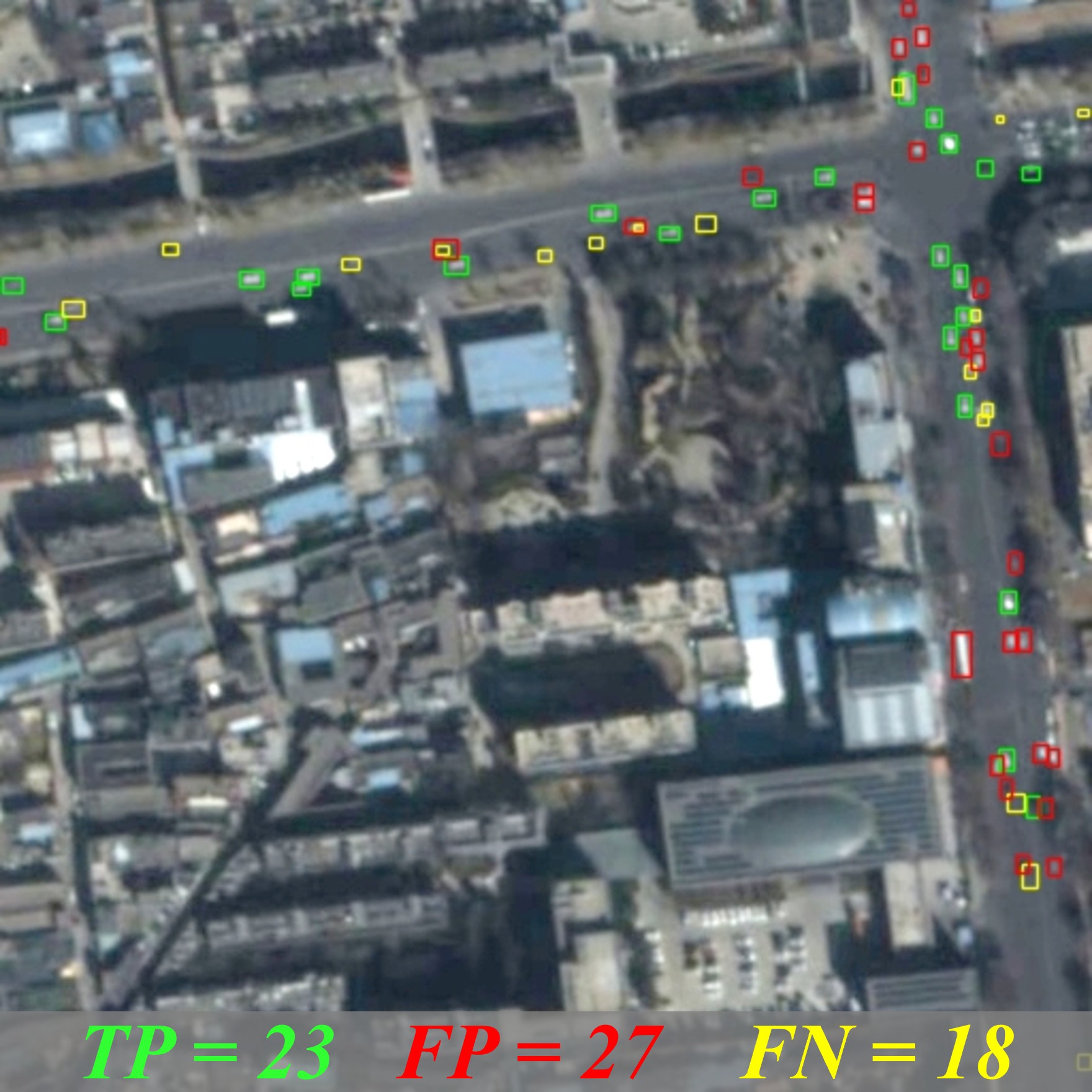} &
\includegraphics[width=0.19\textwidth,height=2.94cm,keepaspectratio]{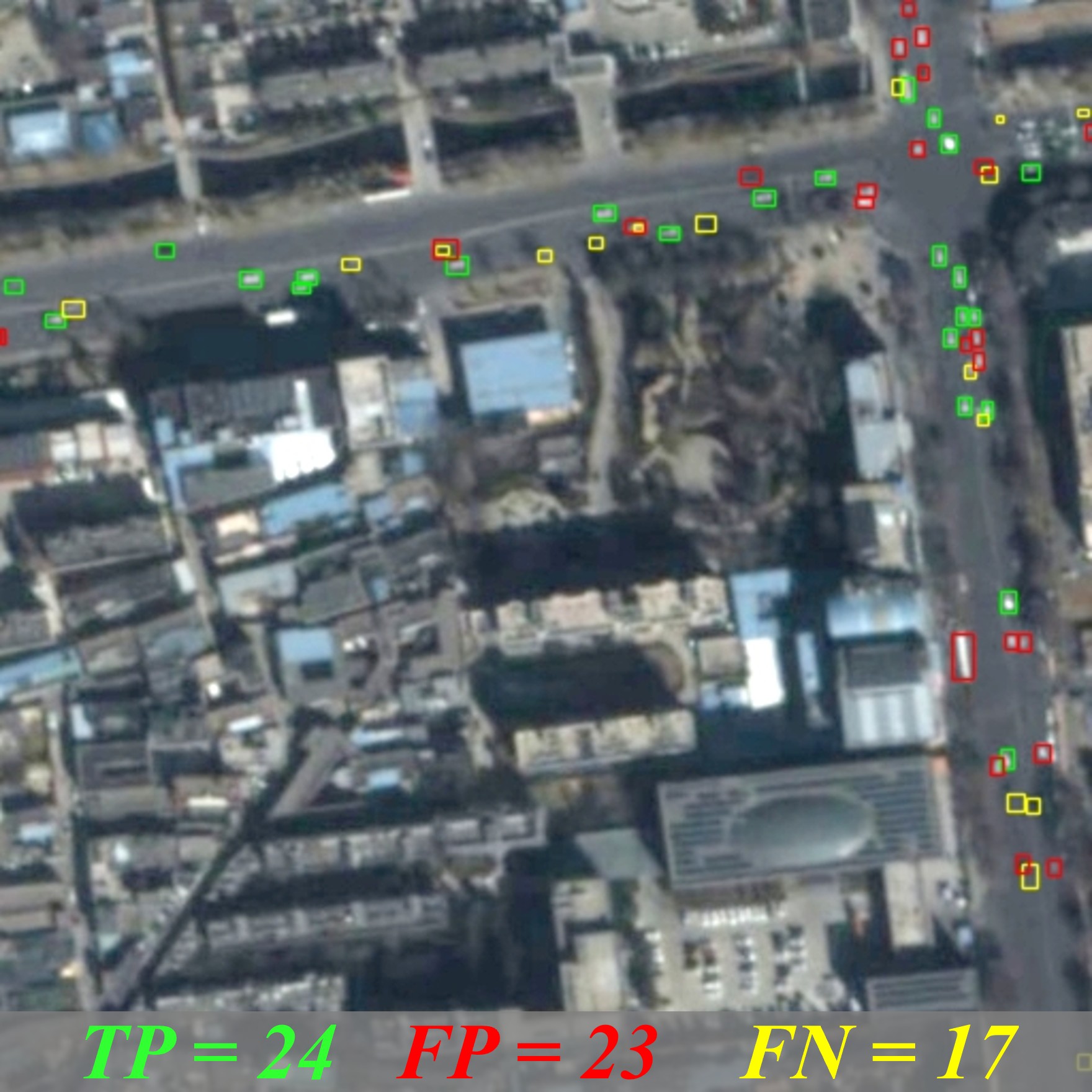} &
\includegraphics[width=0.19\textwidth,height=2.94cm,keepaspectratio]{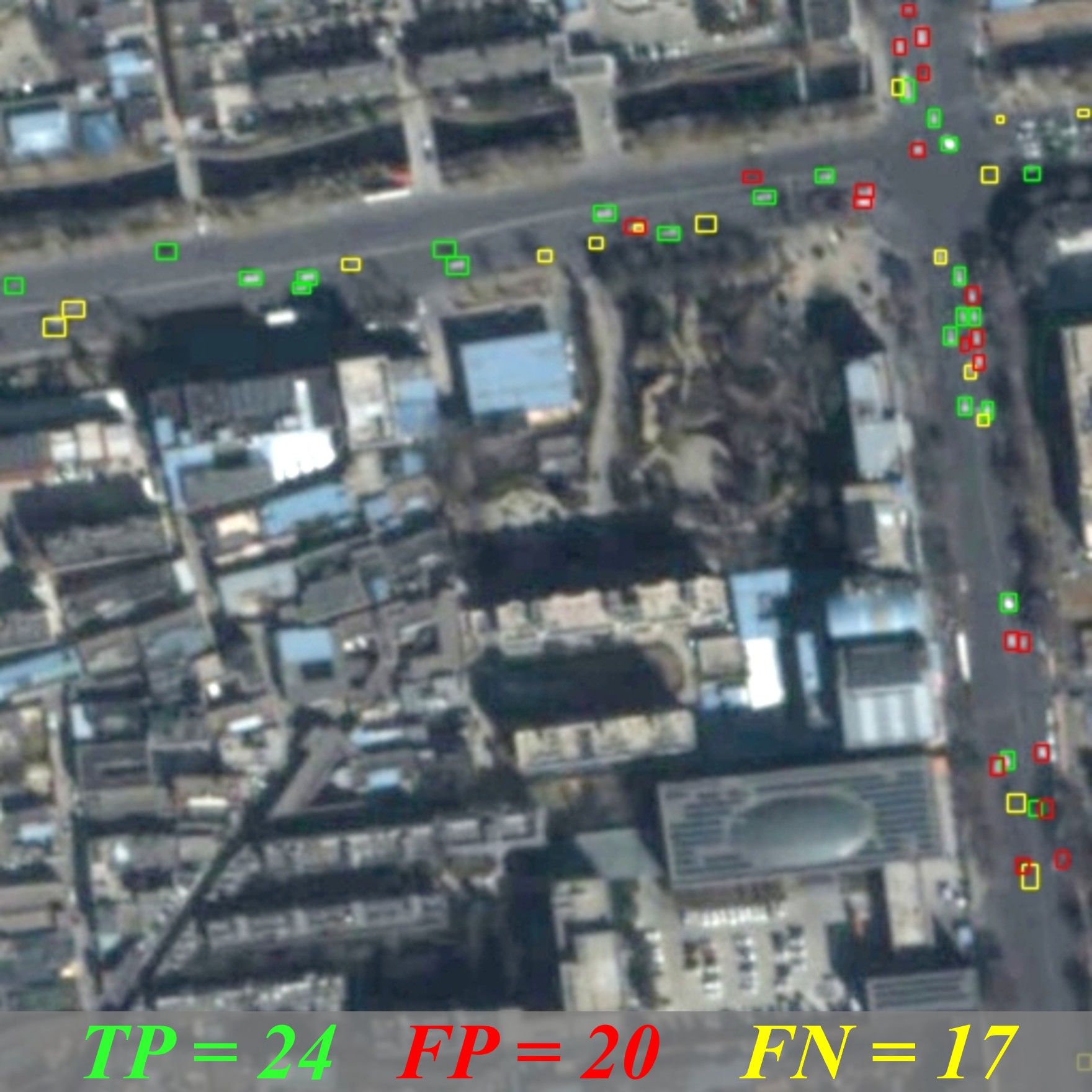} &
\includegraphics[width=0.19\textwidth,height=2.94cm,keepaspectratio]{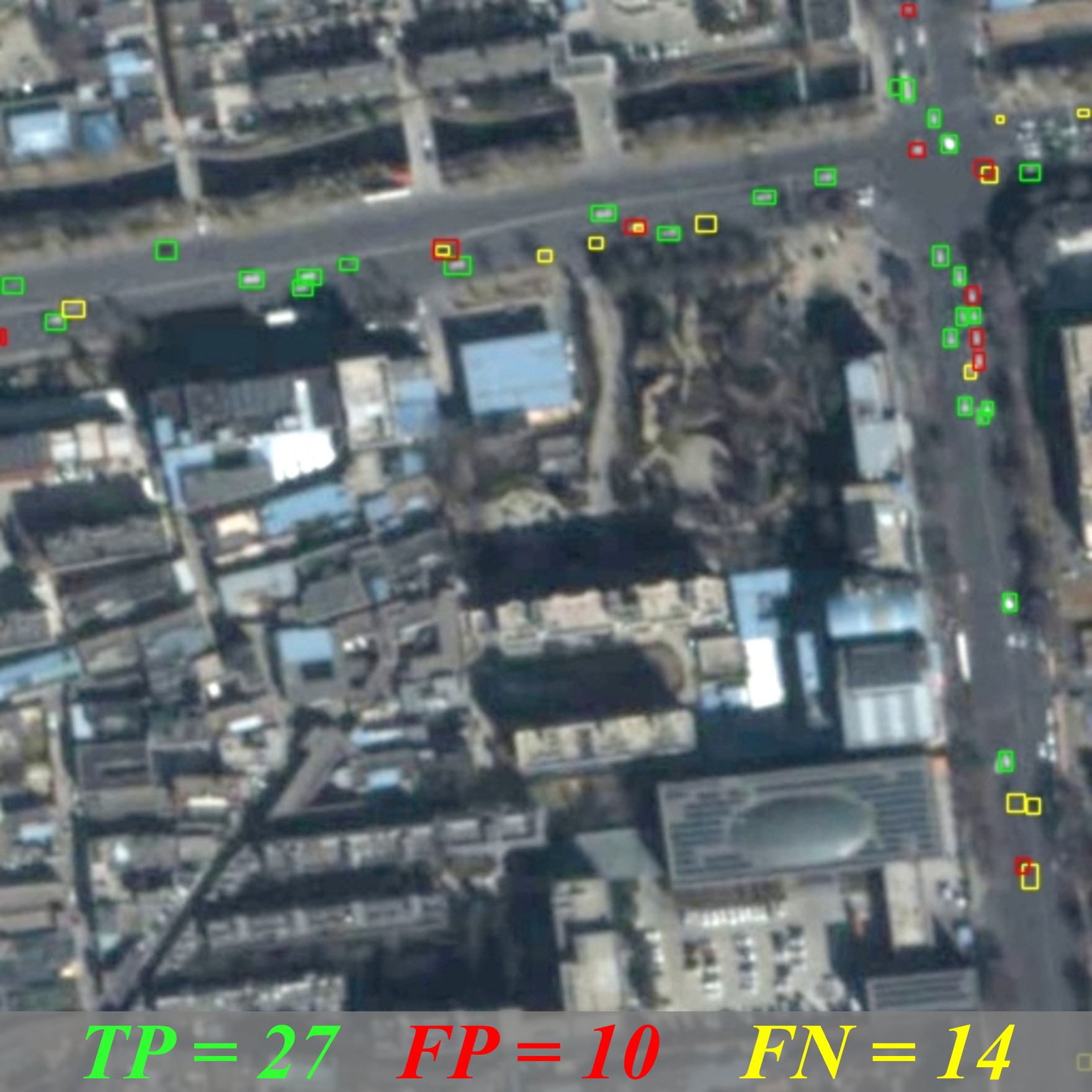} &
\includegraphics[width=0.19\textwidth,height=2.94cm,keepaspectratio]{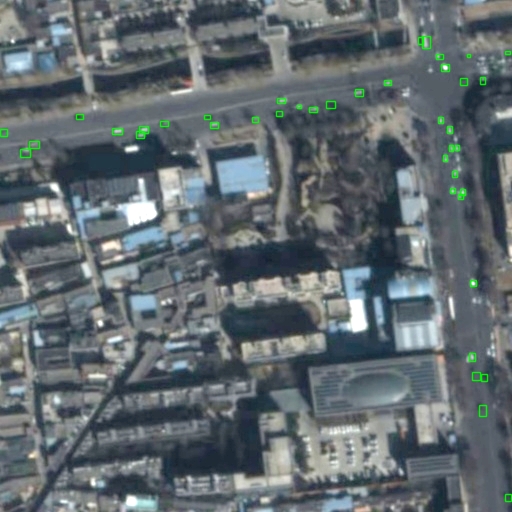} & \\
 FairMOT~\cite{Zhang2021FairMOT} & DSFNet~\cite{Xiao2022DSFNet} &  MP2Net~\cite{Zhao2024MP2Net} & PIFTrack~\cite{Chen2025PIFTrack} & \textbf{DeTracker (Ours)}& Ground truth \\
    \end{tabular}}
   \caption{Examples of detection results for vehicles moving in different directions. From the first line to the third line are frames 50, 120, and 189 respectively, and vehicles undergo lateral and longitudinal motion. Green boxes denote true positives and correct predictions, red boxes indicate false positives, and yellow boxes represent false negatives.}
\label{fig:dingxing}
\end{figure*}

\tabcolsep=0.5pt
\begin{figure*}[htb]
    \centering
\small{
        \begin{tabular}{ccccccc}
           
\includegraphics[width=0.19\textwidth,height=3.4cm,keepaspectratio]{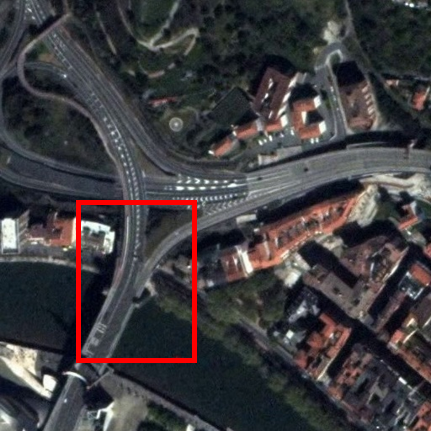} &
\includegraphics[width=0.19\textwidth,height=3.4cm,keepaspectratio]{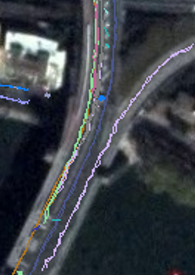} &
\includegraphics[width=0.19\textwidth,height=3.4cm,keepaspectratio]{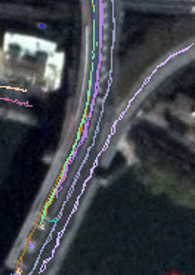} &
\includegraphics[width=0.19\textwidth,height=3.4cm,keepaspectratio]{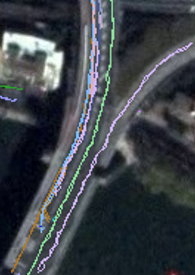} &
\includegraphics[width=0.19\textwidth,height=3.4cm,keepaspectratio]{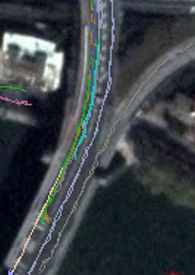} &
\includegraphics[width=0.19\textwidth,height=3.4cm,keepaspectratio]{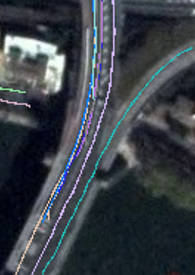} & 
\includegraphics[width=0.19\textwidth,height=3.4cm,keepaspectratio]{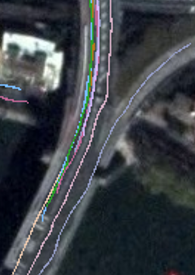} \\
           
\includegraphics[width=0.19\textwidth,height=3.4cm,keepaspectratio]{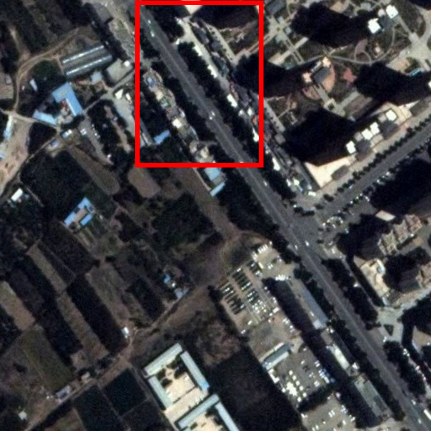} &
\includegraphics[width=0.19\textwidth,height=3.4cm,keepaspectratio]{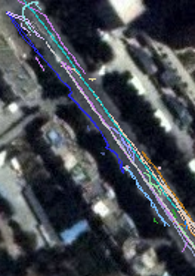} &
\includegraphics[width=0.19\textwidth,height=3.4cm,keepaspectratio]{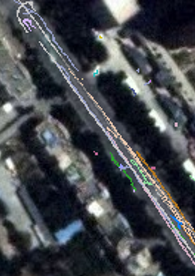} &
\includegraphics[width=0.19\textwidth,height=3.4cm,keepaspectratio]{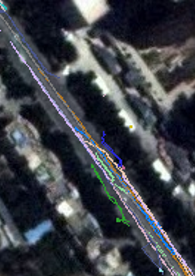} &
\includegraphics[width=0.19\textwidth,height=3.4cm,keepaspectratio]{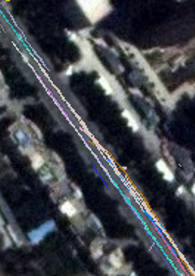} &
\includegraphics[width=0.19\textwidth,height=3.4cm,keepaspectratio]{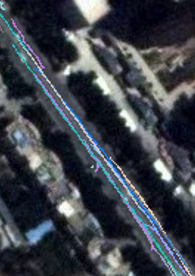} & 
\includegraphics[width=0.19\textwidth,height=3.4cm,keepaspectratio]{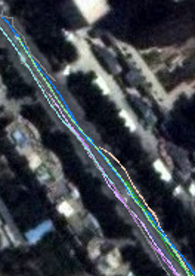} \\
Input frame & FairMOT~\cite{Zhang2021FairMOT} & DSFNet~\cite{Xiao2022DSFNet} &  MP2Net~\cite{Zhao2024MP2Net} & PIFTrack~\cite{Chen2025PIFTrack} & \makebox[0.12\textwidth]{\textbf{DeTracker (Ours)}}& Ground truth \\
    \end{tabular}}
   \caption{Examples of tracking trajectories, where the motion trajectory of each object is shown in a distinct color within the same frame.}
\label{fig:track}
\end{figure*}

\subsubsection{Qualitative Analysis}
We visualize detection results on randomized camera-shake video sequences with diverse vehicle motion directions for different methods, as illustrated in Fig.~\ref{fig:dingxing}.
Under complex backgrounds and severe platform jitter, most baseline methods suffer from frequent false positives and missed detections, particularly in regions with strong illumination changes, shadows, and road boundaries, where bounding boxes often drift or misalign. 
In contrast, DeTracker accurately localizes vehicle targets across consecutive frames, reducing both false positives and missed detections.

Fig.~\ref{fig:track} further visualizes trajectory predictions from different trackers, where all trajectories are projected onto the last frame of each sequence and color coded by vehicle identity. 
As shown in the first row, DeTracker generates smooth and continuous trajectories that closely follow the ground truth paths, remain well aligned with road geometry, and exhibit minimal jitter.  
In contrast, most baselines suffer from notable trajectory fluctuations at turning regions, with some trajectories drifting along background structures. 
In the second row, baseline methods produce fragmented or discontinuous trajectories, indicating identity switches and unstable temporal association. Notably, MP2Net and DSFNet exhibit clear trajectory offsets and breaks, whereas DeTracker consistently predicts complete and coherent trajectories. 
We attribute this performance gain to the GLMD module, which decouples global background motion and enables feature-level alignment, and the TDFP module, which strengthens cross-frame dependency modeling, thereby preserving trajectory coherence and identity consistency over long-term tracking.

\subsection{Ablation Studies} 
\subsubsection{Effectiveness of GLMD}
The GLMD module is designed to decouple satellite platform motion from object motion through global alignment and local refinement across adjacent frames.
As shown in Table~\ref{tab:ABLATION}, incorporating GLMD into the baseline model leads to a substantial performance gain, improving MOTA by 12.0\% (from 40.5\% to 52.5\%) and IDF1 by 1.1\% (from 64.9\% to 66.0\%). Notably, the reduction in IDs and the corresponding improvement in IDF1 are primarily attributed to the stable feature alignment and precise target localization capabilities provided by GLMD. By effectively decoupling the complex platform motion from the target intrinsic motion, GLMD significantly mitigates tracking failures and ID switches caused by background drift.

\begin{table}[t]
  \centering
  \caption{Ablation study on the \emph{SDM-Car-SU-1} dataset}
  \label{tab:ABLATION}
  \begin{tabular*}{\columnwidth}{@{\extracolsep{\fill}}cccccccc}
    \toprule
    Baseline & GLMD & TDFP& MOTA$\uparrow$ & IDF1$\uparrow$&  IDs$\downarrow$ & FP$\downarrow$ & FN$\downarrow$ \\
    \midrule
    \checkmark &  &  & 40.5\% & 64.9\% & 3103 & 56784 & 63141 \\
    \checkmark & \checkmark &  & 52.5\% & 66.0\%& 2932 & 34937 & 40085  \\
    \checkmark &  &\checkmark  & 57.6\% & 67.8\% & 2824 & 41122 & 49469 \\
     \checkmark & \checkmark & \checkmark & 61.1\% & 68.7\%  & 1863 & 23065 & 33046  \\ 
    \bottomrule
  \end{tabular*}
\end{table}

\begin{table}[t]
  \centering
  \caption{Ablation study for GLMD on the \emph{SDM-Car-SU-3} dataset}
  \label{tab:ABLATIONGLMD}
  \begin{tabular*}{\columnwidth}{@{\extracolsep{\fill}}cccccccc}
    \toprule
    Baseline & GA & LR& SIFT& Cosine Similarity& MOTA$\uparrow$ & IDF1$\uparrow$&  IDs$\downarrow$  \\
    \midrule
    \checkmark &  &  & & 0.981 &33.0\% & 47.1\% & 5535  \\
    \checkmark & \checkmark & &  &0.985 & 41.5\% & 57.1\%&  4288 \\
    \checkmark &  & &\checkmark & 0.992 & 35.3\% & 55.8\% & 5548  \\
     \checkmark & \checkmark & \checkmark & & 0.988 & 52.4\% & 63.8\%  & 3231  \\ 
    \bottomrule
  \end{tabular*}
\end{table}

\begin{figure*}[!htb]
    \centering
    \setlength{\tabcolsep}{1.5pt} 
    {\small
    \begin{tabular}{cccc}
        \includegraphics[height=4cm,keepaspectratio]{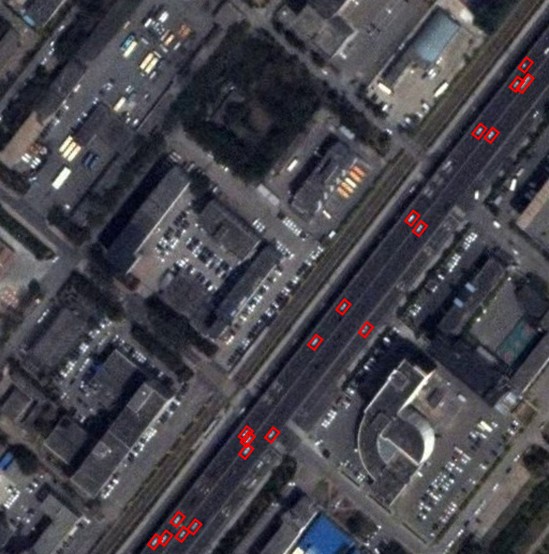} &
        \includegraphics[height=4cm,keepaspectratio]{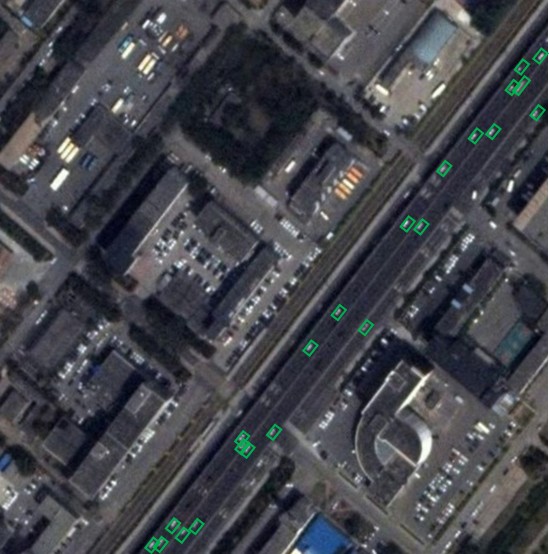} &
        \includegraphics[height=4cm,keepaspectratio]{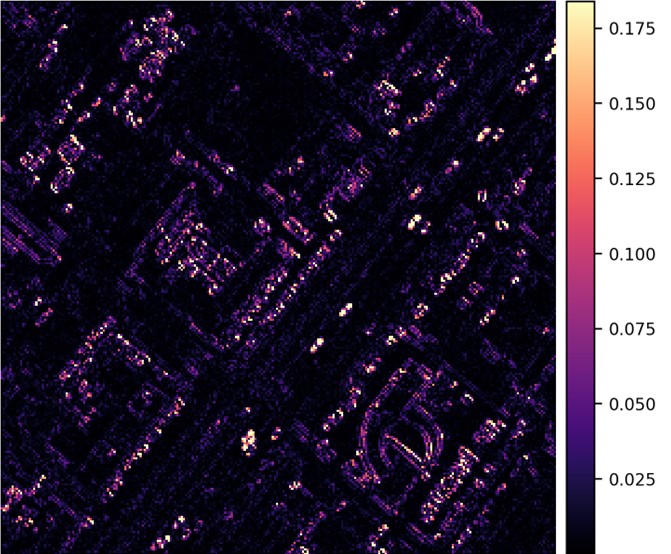} &
        \includegraphics[height=4cm,keepaspectratio]{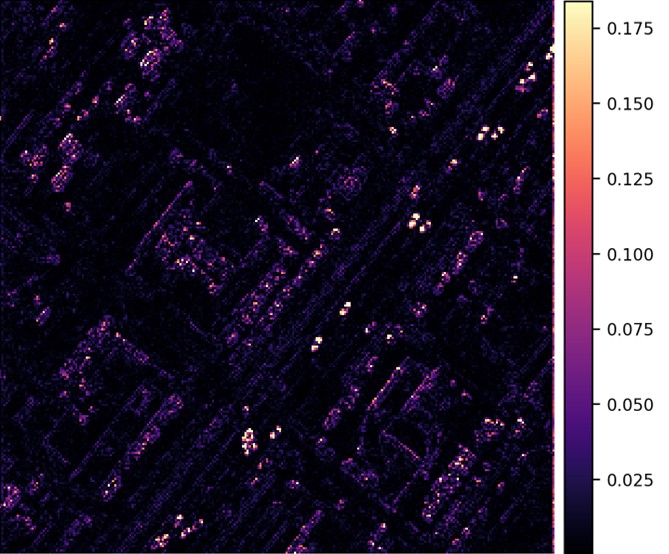} \\

        Frame\#005 &
        Frame\#006 &
        Before GLMD &
        After GLMD \\
    \end{tabular}}

    \caption{Visualization of the GLMD effect. Before alignment, the feature difference map between two consecutive frames contains many highlighted regions, indicating large discrepancies.
After alignment, the overall brightness of the difference map decreases, suggesting reduced inter-frame feature variance.}
    \label{fig:vsialign}
\end{figure*}

Fig.~\ref{fig:vsialign} provides qualitative evidence of GLMD's effectiveness. The brightly colored regions denote larger feature discrepancies between the current frame (frame006) and the reference frame (frame005). After applying GLMD, the differences between the aligned features and those of the reference frame become notably smaller, especially in background areas, and the overall error distribution is more concentrated. The alignment-improvement map further localizes areas where GLMD brings tangible benefits: the dominant dark-purple response across most of the scene reflects a consistent global enhancement in feature-space alignment.

To further verify the contribution of individual components within the GLMD module, 
Table~\ref{tab:ABLATIONGLMD} presents the ablation results on the \emph{SDM-Car-SU} dataset under the most challenging setting, U3 (denoted as \emph{SDM-Car-SU-3}).
It can be observed that the global alignment branch alone leads to an 8.5\% improvement in MOTA, 
indicating that global feature alignment effectively suppresses large-scale drifts caused by platform motion. 
When a traditional SIFT-based method~\cite{SIFT2017} is introduced, feature-level similarity increases, but it yields a lower MOTA than the learning-based global alignment branch. This suggests that handcrafted keypoint-based methods are less reliable for aligning tiny objects, which are prone to misalignment under severe jitter and low-resolution conditions. 
By jointly integrating both global alignment and local refinement, the MOTA and IDF1 scores further increase to 52.4\% and 63.8\%, respectively, demonstrating that local refinement effectively captures small-scale local motions
and helps cleanly separate global and local motion patterns.

\subsubsection{Effectiveness of TDFP}
The TDFP module explicitly enhances object features through temporal feature fusion.
As shown in Table~\ref{tab:ABLATION}, incorporating TDFP leads to notable gains in the primary evaluation metrics, improving MOTA by 17.1\% (from 40.5\% to 57.6\%) and IDF1 by 2.9\% (from 64.9\% to 67.8\%), indicating a substantial improvement in long-term tracking success. These improvements are attributed to the implicit temporal consistency modeling introduced by the TDFP module at the feature level. By utilizing ConvGRU to recursively propagate and aggregate spatio-temporal features across consecutive frames, the TDFP establishes a cross-frame memory mechanism, thereby maintaining the temporal continuity and discriminative stability of targets within the feature space.

\begin{figure}[t]
    \centering
    \setlength{\tabcolsep}{0.5pt} 

    {\small
    \begin{tabular}{cccc}
        \includegraphics[height=2.2cm,keepaspectratio]{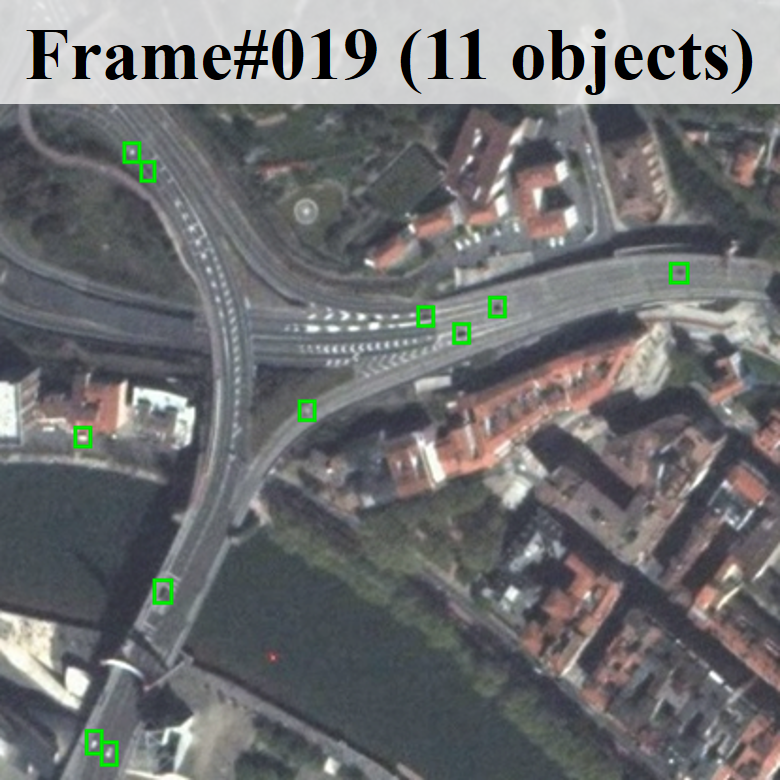} &
        \includegraphics[height=2.2cm,keepaspectratio]{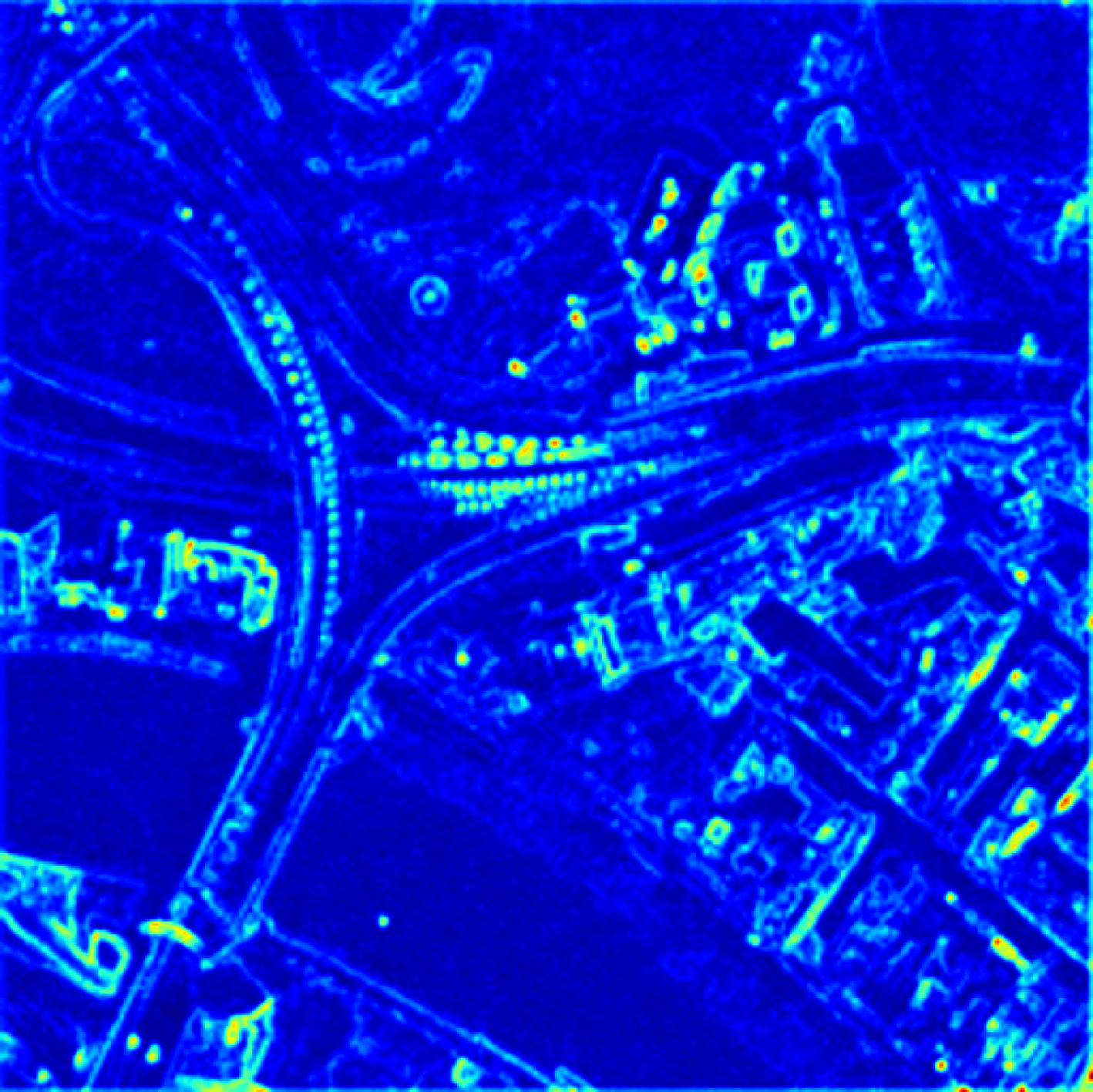} &
        \includegraphics[height=2.2cm,keepaspectratio]{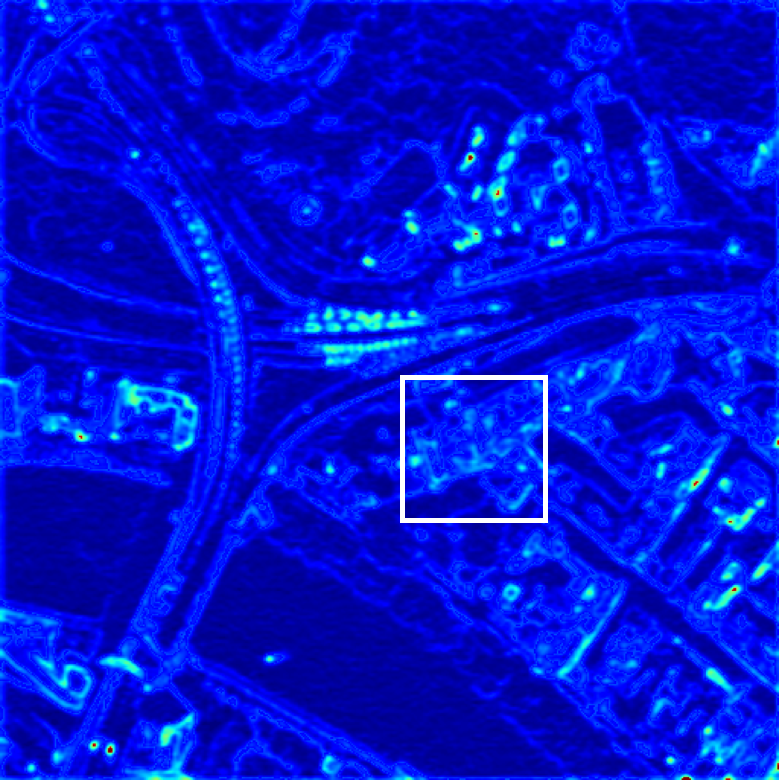} &
        \includegraphics[height=2.2cm,keepaspectratio]{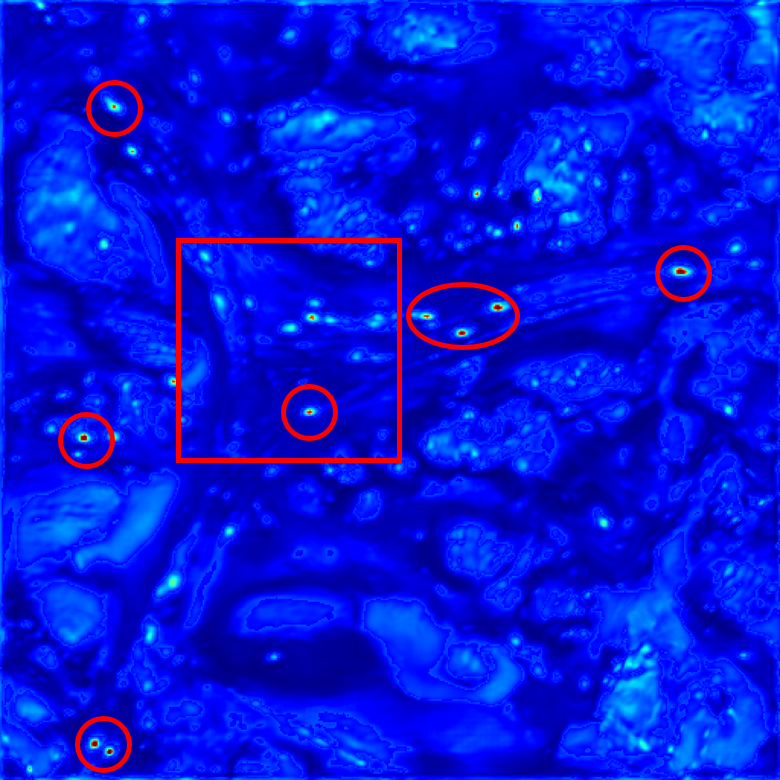} \\[-2pt]
        \includegraphics[height=2.2cm,keepaspectratio]{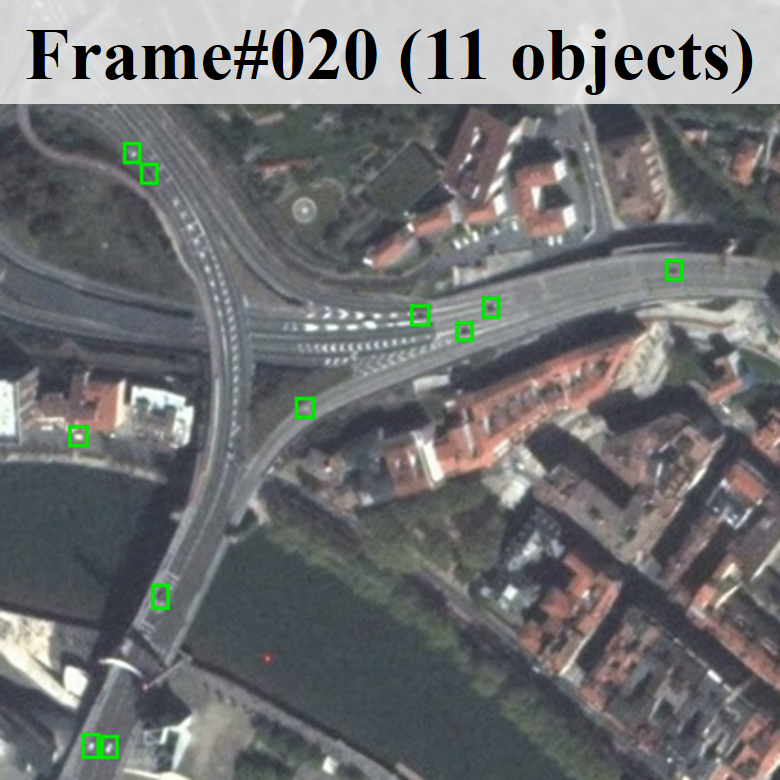} &
        \includegraphics[height=2.2cm,keepaspectratio]{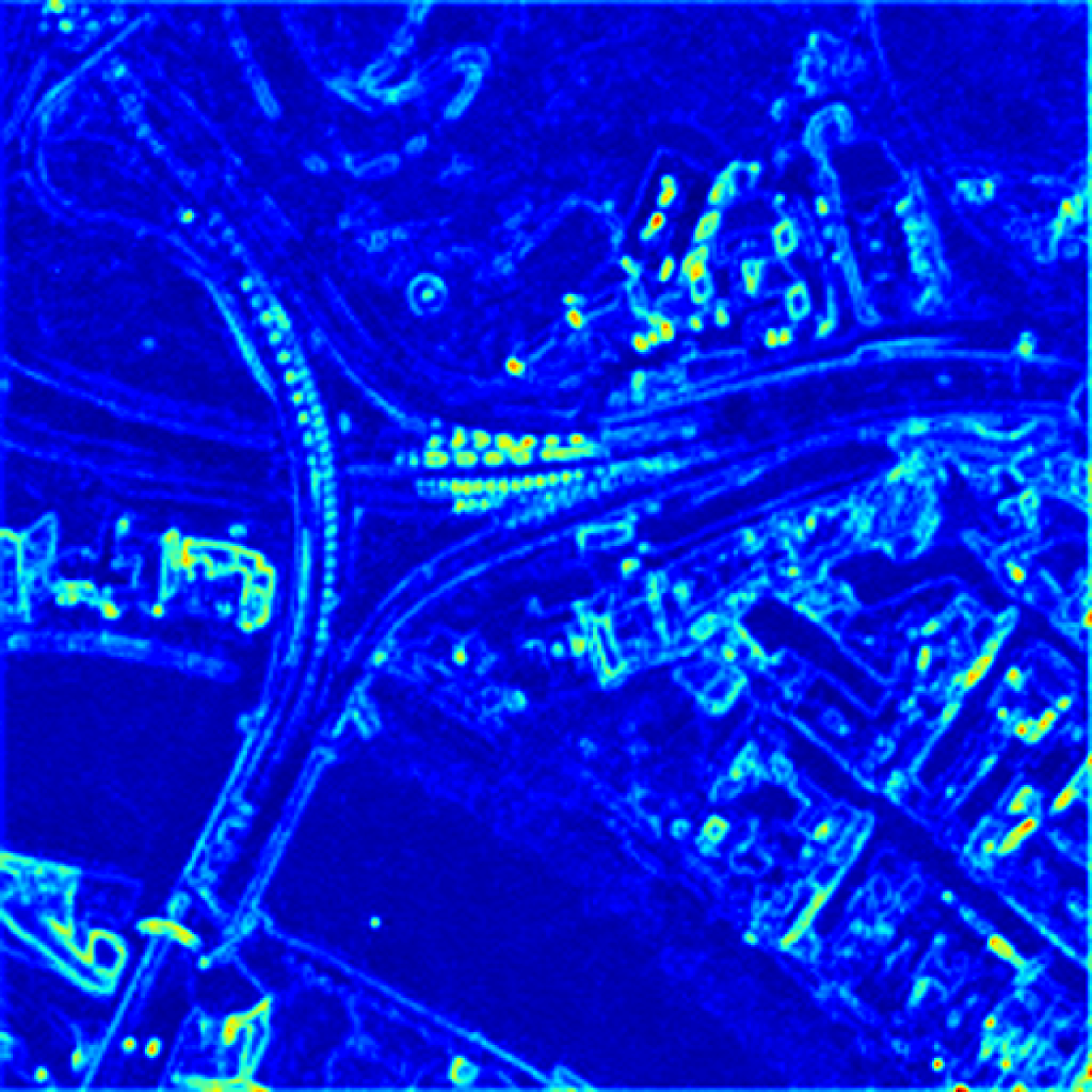} &
        \includegraphics[height=2.2cm,keepaspectratio]{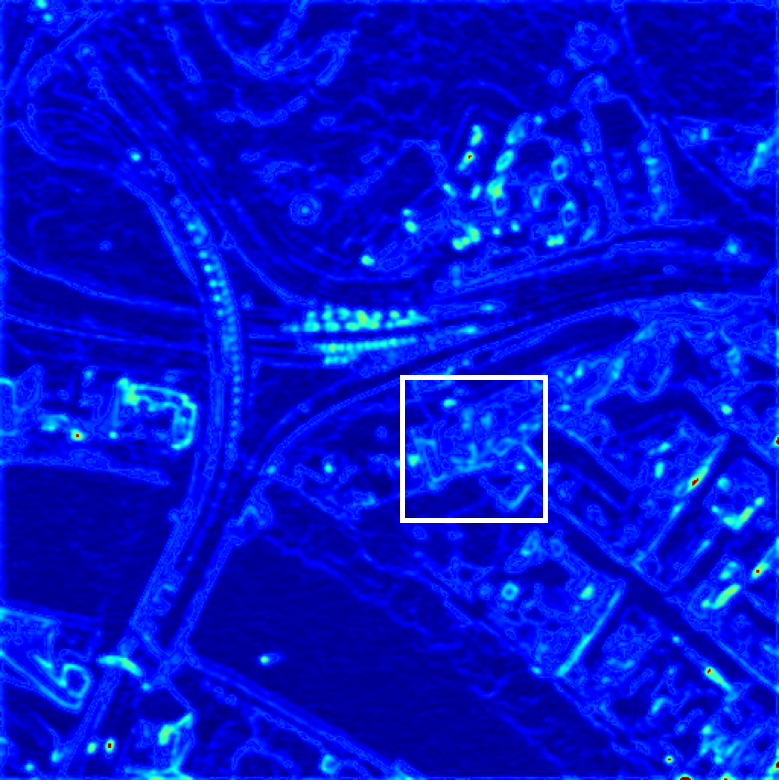} &
        \includegraphics[height=2.2cm,keepaspectratio]{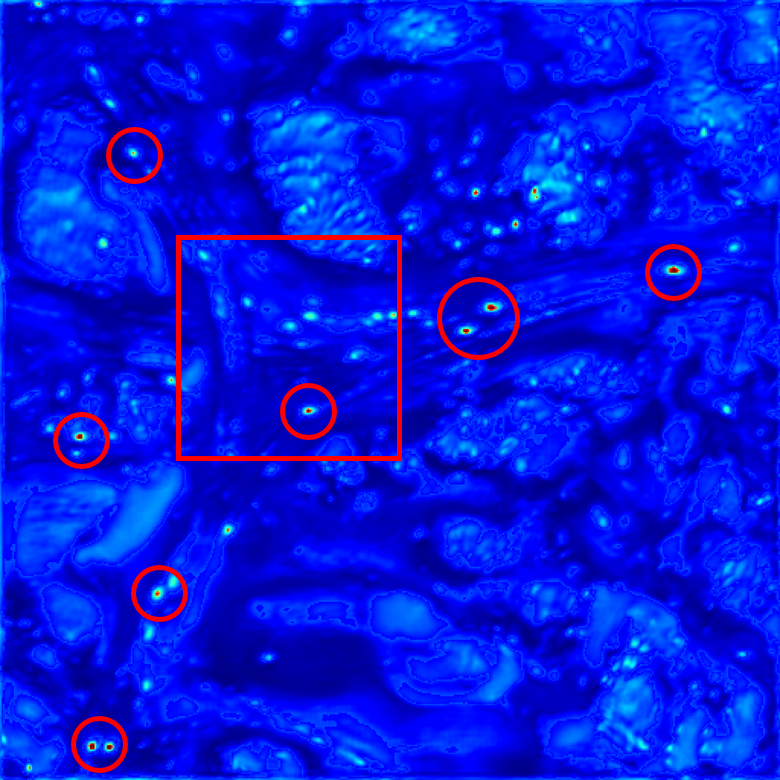} \\
        Image &
        Baseline &
        With GLMD &
        With TDFP \\
    \end{tabular}
    }

    \caption{Feature map visualization of adjacent frames after incorporating different modules. Compared with the baseline, the GLMD module effectively corrects background misalignment and reduces static interference (indicated by the white boxes) through global feature alignment. Building on this, the further integration of the TDFP module improves the feature representation of tiny moving objects (indicated by red circles) while filtering out background distractions, such as lane markings (indicated by red boxes).}
    \label{fig:Ab_td}
\end{figure}

Fig.~\ref{fig:Ab_td} further illustrates its effect. Compared with the baseline, TDFP yields more discriminative and stable feature representations for tiny objects, significantly improving cross-frame consistency. As a result, the model remains robust to inter-frame feature fluctuations and preserves reliable object recognition throughout the tracking process.

\subsubsection{Hyperparameter Analysis}
Table~\ref{tab:k} presents the impact of the input frame number $k$ and loss weighting parameters $\alpha$ and $\beta$ on the overall performance. It can be observed that both MOTA and IDF1 exhibit an overall upward trend as $k$ increases, with the best performance obtained at $k=5$, indicating that a moderately enlarged temporal window effectively enhances cross-frame feature consistency and association stability. 
However, when $k=6$, the longer-term dependencies introduce redundant background and jitter information, which in turn weakens the quality of motion modeling.

Regarding the loss weight design, we conduct a systematic sensitivity analysis on the classification loss weight $\alpha$ and the bounding-box regression loss weight $\beta$. Given that the overall loss function includes the warp alignment loss, we perform a comprehensive evaluation within the ranges of $\alpha \in \{0.2, 0.3\}$ and $\beta \in \{0.1, 0.2, 0.3\}$ to ensure a balance among the various loss components. As shown in Table~\ref{tab:k}, when $\alpha$ is fixed, a smaller $\beta$ leads to insufficient localization constraints, thereby increasing false positives; conversely, a larger $\beta$ restricts the optimization space for the warp alignment loss, negatively affecting the overall tracking performance. Overall, the optimal performance is achieved when $\alpha=0.3$ and $\beta=0.2$.

\begin{table}[t]
  \centering
  \caption{Comparison on \emph{SDM-Car-SU} test set with different values of the hyperparameters}
  \label{tab:k}
  \begin{tabular*}{\columnwidth}{@{\extracolsep{\fill}}cccccc}
    \toprule
     & MOTA$\uparrow$ & IDF1$\uparrow$ & IDs$\downarrow$ & FP$\downarrow$ & FN$\downarrow$ \\
    \midrule
    $k$=3  & 54.6\% & 62.8\% & 1988 & 33128 & 38177\\
    $k$=4 & 57.7\% & 66.7\% & 1896 & 30058 & 34570\\ 
    $k$=5 &  61.1\% & 68.7\% & 1863 & 23065 & 33046\\ 
    $k$=6 & 58.5\% & 67.9\% &1904 & 23974 & 34734\\  
    \midrule
    $\alpha$=0.2, $\beta$=0.1 & 58.8\% & 62.8\% & 1914 & 25179 & 34301 \\
    $\alpha$=0.2, $\beta$=0.2 & 58.9\% & 63.4\% & 1909 & 24858 & 34480 \\
    $\alpha$=0.2, $\beta$=0.3 & 58.7\% & 62.3\% & 1890 & 25486 & 34159 \\
    $\alpha$=0.3, $\beta$=0.1 & 58.3\% & 61.7\% & 1928 & 25766 & 34447 \\
    $\alpha$=0.3, $\beta$=0.2 & 61.1\% & 68.7\% & 1863 & 23065 & 33046 \\
    $\alpha$=0.3, $\beta$=0.3 & 60.1\% & 66.3\% & 1886 & 23741 & 33833 \\
    \bottomrule
  \end{tabular*}
\end{table}

\begin{table}[!htb]
  \centering
  \caption{Efficiency comparison on real unstabilized satellite videos}
   \label{tab:FPS}
  \begin{tabular*}{0.8\columnwidth}{@{\extracolsep{\fill}}lccc}
    \toprule
     Method & Params (M) & FLOPs (G) & FPS  \\
    \midrule
    FairMOT~\cite{Zhang2021FairMOT} &  20.5 & 61.2&13.6 \\
    DSFNet~\cite{Xiao2022DSFNet}  & 6.1 & 49.7 & 3.4\\
    MP2Net~\cite{Zhao2024MP2Net} & 7.2 & 80.5 & 2.3 \\  
    PIFTrack~\cite{Chen2025PIFTrack}& 32.4 & 57.4 &10.8 \\ 
     \midrule
     \textbf{DeTracker (Ours)}  & 8.71 & 79.8 & 4.86 \\ 
    \bottomrule
  \end{tabular*}
\end{table}

\subsubsection{Computational Complexity}
As shown in Table~\ref{tab:FPS}, DeTracker contains 8.71M parameters with a computational cost of 79.8 GFLOPs and an inference speed of approximately 4.86 FPS, while real-time processing is generally considered to require around 5–10 FPS. The GLMD and TDFP modules leverage cross-frame motion information to enhance tracking accuracy and involve corresponding computational overhead.

\section{Conclusion}
This paper presents DeTracker, a joint-detection-and-tracking (JDT) framework tailored for unstabilized satellite videos, aiming to achieve robust multi-object tracking in complex dynamic scenes without relying on any stabilization preprocessing. 
To explicitly decouple satellite platform motion from true object motion, we introduce the GLMD module, which performs two-stage motion modeling via global alignment and local refinement. This design effectively compensates for large-scale background motion while preserving fine-grained object dynamics, leading to more accurate trajectory estimation and improved motion consistency. In addition, the proposed TDFP module enables multi-scale feature fusion and cross-frame dependency modeling, substantially enhancing the continuity and discriminability of tiny-object representations over time. 
Furthermore, we release a new simulated benchmark dataset, \emph{SDM-Car-SU}, which incorporates diverse motion directions and intensities to facilitate systematic evaluation of algorithmic robustness under unstabilized conditions. Extensive experiments on both simulated and real satellite videos demonstrate that DeTracker consistently outperforms state-of-the-art methods, validating its effectiveness for practical satellite video multi-object tracking.

\bibliographystyle{IEEEtran}  
\bibliography{TGRS2025}

\end{document}